\documentclass{nle}
\usepackage{graphicx}
\usepackage[utf8]{inputenc}	
\usepackage{lipsum}
\usepackage{soul}
\usepackage{listings}
\usepackage{amsmath}
\usepackage{color}
\usepackage{subcaption}
\usepackage{soul}
\usepackage[dvipsnames]{xcolor}
\usepackage{multirow}
\usepackage{url}
\usepackage{longtable}
\usepackage{natbib}
\usepackage[british]{babel}
\usepackage{hyperref}
\usepackage{tcolorbox}
\newcounter{example}
\definecolor{blackish1}{rgb}{0.3, 0.3, 0.3}
\definecolor{blackish2}{rgb}{0.1, 0.1, 0.1}

\usepackage{color}
\def\paperDraft{}

\ifdefined\paperDraft
 \def\spcomment#1{{\color{blue}[Senja: \textit{#1}]}}
 
\else
 \def\spcomment#1{}
 
\fi

\begin{document}

\newcommand\doc[1]{
   \leavevmode\par
   \stepcounter{example}
   \noindent
   \color{blackish1}
   \textbf{Document \theexample:} \\  #1\par }
   
\newcommand\pred[1]{
    \leavevmode\par\noindent                  
    \color{blackish2}
   {\leftskip10pt
     \textbf{Predicted keywords: }#1\par}}
     
\newcommand\true[1]{
    \leavevmode\par\noindent                  
    \color{blackish2}
   {\leftskip10pt
     \textbf{True keywords: }#1\par}}

\label{firstpage}

\lefttitle{}
\righttitle{Natural Language Engineering}

\papertitle{Article}

\jnlPage{1}{00}
\jnlDoiYr{2020}
\doival{10.1017/xxxxx}

\title{TNT-KID: Transformer-based Neural Tagger for Keyword Identification}

\titlerunning{TNT-KID: Transformer-based Neural Tagger for Keyword Identification}
\begin{authgrp}
\author{Matej Martinc$^{*1,2}$, Blaž Škrlj$^{1,2}$, Senja Pollak$^{1}$}
\affiliation{$^{1}$Jo\v{z}ef Stefan Institute,\\ Department of Knowledge Technologies, Jamova 39, 1000 Ljubljana, Slovenia}
\affiliation{$^2$Jo\v{z}ef Stefan International Postgraduate School,\\
Department of Knowledge Technologies, Jamova 39, 1000 Ljubljana, Slovenia\\
\email{$^{*}$Corresponding author. Email:matej.martinc@ijs.si}}
\end{authgrp}

\authorrunning{M. Martinc et al.} 


\begin{tcolorbox}
\noindent The  final reviewed  publication  was published in the Natural Language Engineering Journal and is available online at \url{https://doi.org/10.1017/S1351324921000127}.
\end{tcolorbox}

\begin{abstract}
With growing amounts of available textual data, development of algorithms capable of automatic analysis, categorization and summarization of these data has become a necessity. In this research we present a novel algorithm for keyword identification, i.e., an extraction of one or multi-word phrases representing key aspects of a given document, called Transformer-based Neural Tagger for Keyword IDentification (TNT-KID). By adapting the transformer architecture for a specific task at hand and leveraging language model pretraining on a domain specific corpus, the model is capable of overcoming deficiencies of both supervised and unsupervised state-of-the-art approaches to keyword extraction by offering competitive and robust performance on a variety of different datasets while requiring only a fraction of manually labeled data required by the best performing systems. This study also offers thorough error analysis with valuable insights into the inner workings of the model and an ablation study measuring the influence of specific components of the keyword identification workflow on the overall performance. 
\end{abstract}

\maketitle
\section{Introduction}
\label{intro}

With the exponential growth in amount of available textual resources, organization, categorization and summarization of these data presents a challenge, the extent of which becomes even more apparent when it is taken into the account that a majority of these resources do not contain any adequate meta information. Manual categorization and tagging of documents is unfeasible due to a large amount of data, therefore development of algorithms capable of tackling these tasks automatically and efficiently has become a necessity \citep{firoozeh2020keyword}.   

One of the crucial tasks for organization of textual resources is keyword identification, which deals with automatic extraction of words that represent crucial semantic aspects of the text and summarize its content. First automated solutions to keyword extraction have been proposed more than a decade ago \citep{witten2005kea,mihalcea2004textrank} and the task is currently again gaining traction, with several new algorithms proposed in the recent years. Novel unsupervised approaches, such as RaKUn \citep{vskrlj2019rakun} and YAKE \citep{campos2018yake}, work fairly well and have some advantages over supervised approaches, as they are language and genre independent, do not require any training and are computationally undemanding. On the other hand, they also have a couple of crucial deficiencies:

\begin{itemize}
    \item Term frequency - inverse document frequency (TfIdf) and graph based features, such as PageRank, used by these systems to detect the importance of each word in the document, are based only on simple statistics like word occurrence and co-occurrence, and are therefore unable to grasp the entire semantic information of the text.
    
    \item Since these systems cannot be trained, they can not be adapted to the specifics of the syntax, semantics, content, genre and keyword assignment regime of a specific text (e.g., a variance in a number of keywords).
\end{itemize}

These deficiencies result in a much worse performance when compared to the state-of-the-art supervised algorithms (see Table \ref{tbl:results}), which have a direct access to the gold standard keyword set for each text during the training phase, enabling more efficient adaptation. The newest supervised neural algorithms \citep{meng2019does, yuan2019one} therefore achieve excellent performance under satisfactory training conditions and can model semantic relations much more efficiently than algorithms based on simpler word frequency statistics. On the other hand, these algorithms are resource demanding, require vast amount of domain specific data for training and can therefore not be used in domains and languages that lack manually labeled resources of sufficient size.     

In this research we propose Transformer-based Neural Tagger for Keyword IDentification (TNT-KID)\footnote{Code is available under the MIT license at \url{https://gitlab.com/matej.martinc/tnt_kid/}.} that is capable of overcoming the aforementioned deficiencies of supervised and unsupervised approaches. We show that, while requiring only a fraction of manually labeled data required by other neural approaches, the proposed approach achieves performance comparable to the state-of-the-art supervised approaches on test sets for which a lot of manually labeled training data is available. On the other hand, if training data that is sufficiently similar to the test data is scarce, our model outperforms state-of-the-art approaches by a large margin. This is achieved by leveraging the transfer learning technique, where a keyword tagger is first trained in an unsupervised way as a language model on a large corpus and then fine-tuned on a (usually) small-sized corpus with manually labeled keywords. By conducting experiments on two different domains, computer science articles and news, we show that the language model pretraining allows the algorithm to successfully adapt to a specific domain and grasp the semantic information of the text, which drastically reduces the needed amount of labeled data for training the keyword detector. 

The transfer learning technique \citep{Peters:2018,howard2018universal}, which has recently become a well established procedure in the field of natural language processing (NLP), in a large majority of cases relies on very large unlabeled textual resources used for language model pretraining. For example, a well known English BERT model \citep{devlin2018bert} was pretrained on the Google Books Corpus \citep{goldberg2013dataset} (800 million tokens) and Wikipedia (2,500 million tokens). On the other hand, we show that smaller unlabeled domain specific corpora (87 million tokens for computer science and 232 million tokens for news domain) can be successfully used for unsupervised pretraining, which makes the proposed approach easily transferable to languages with less textual resources and also makes training more feasible in terms of time and computer resources available.     

Unlike most other proposed state-of-the-art neural keyword extractors \citep{meng2017deep, meng2019does, yuan2019one}, we do not employ recurrent neural networks but instead opt for a transformer architecture \citep{vaswani2017attention}, which has not been widely employed for the task at hand. In fact, the study by \cite{sahrawat2019keyphrase} is the only study we are aware of that employs transformers for the keyword extraction task. Another difference between our approach and most very recent state-of-the-art approaches from the related work is also task formulation. While \cite{meng2017deep, meng2019does} and \cite{yuan2019one} formulate a keyword extraction task as a sequence-to-sequence generation task, where the classifier is trained to generate an output sequence of keyword tokens step by step according to the input sequence and the previous generated output tokens, we formulate a keyword extraction task as a sequence labeling task, similar as in \cite{gollapalli2017incorporating}, \cite{luan2017scientific} and \cite{sahrawat2019keyphrase}.

Besides presenting a novel keyword extraction procedure, the study also offers an extensive error analysis, in which the visualization of transformer attention heads is used to gain insights into inner workings of the model and in which we pinpoint key factors responsible for the differences in performance of TNT-KID and other state-of-the-art approaches. Finally, this study also offers a systematic evaluation of several building blocks and techniques used in a keyword extraction workflow in a form of an ablation study. Besides determining the extent to which transfer learning affects the performance of the keyword extractor, we also compare two different pretraining objectives, autoregressive language modelling and masked language modelling \citep{devlin2018bert}, and measure the influence of transformer architecture adaptations, a choice of input encoding scheme and the addition of part-of-speech (POS) tags information on the performance of the model.  

The paper is structured as follows. Section~\ref{sec-relatedwork} addresses the related work on keyword identification and covers several supervised and unsupervised approaches to the task at hand. Section~\ref{sec:methodology} describes the methodology of our approach, while in Section \ref{sec:experimental-setting} we present the datasets, conducted experiments and results. Section \ref{sec:error} covers error analysis, Section \ref{sec:ablation} presents the conducted ablation study, while the conclusions and directions for further work are addressed in Section~\ref{sec:conclusion}.

\section{Related work}
\label{sec-relatedwork}

This section overviews selected methods for keyword extraction, supervised in Section \ref{Sec:SupervisedSOTA} and unsupervised in Section \ref{Sec:UnupervisedSOTA}. The related work is somewhat focused on the newest keyword extraction methods, therefore for a more comprehensive survey of slightly older methods, we refer the reader to \citet{hasan2014automatic}.


\subsection{Supervised keyword extraction methods}
\label{Sec:SupervisedSOTA}

Traditional supervised approaches to keyword extraction considered the task as a two step process (the same is true for  unsupervised approaches). First, a number of syntactic and lexical features are used to extract keyword candidates from the text. Secondly, the extracted candidates are ranked according to different heuristics and the top \textit{n} candidates are selected as keywords \citep{yuan2019one}. One of the first supervised approaches to keyword extraction was proposed by \citet{witten2005kea}, whose algorithm named KEA uses only TfIdf and the term's position in the text as features for term identification. These features are fed to the Naive Bayes classifier, which is used to determine for each word or phrase in the text if it is a keyword or not. \citet{medelyan2009maui} managed to build on the KEA approach and proposed the \textit{Maui} algorithm, which also relies on the Naive Bayes classifier for candidate selection but employs additional semantic features, such as e.g., \textit{node degree}, which quantifies the semantic relatedness of a candidate to other candidates, and \textit{Wikipedia-based keyphraseness}, which is the likelihood of a phrase being a link in the Wikipedia.

A more recent supervised approach is a so-called sequence labelling approach to keyword extraction by \cite{gollapalli2017incorporating}, where the idea is to train a keyword tagger using token-based linguistic, syntactic and structural features. The approach relies on a trained Conditional Random Field (CRF) tagger and the authors demonstrated that this approach is capable of working on-par with slightly older state-of-the-art systems that rely on information from the Wikipedia and citation networks, even if only within-document features are used. Another sequence labeling approach proposed by \cite{luan2017scientific} builds a sophisticated neural network by combing an input layer comprising a concatenation of word, character and part-of-speech embeddings, a bidirectional Long Short-Term Memory (BiLSTM) layer and a CRF tagging layer. They also propose a new semi-supervised graph based training regime for training the network. 

Some of the most recent state-of-the-art approaches to keyword detection consider the problem as a sequence-to-sequence generation task. The first research leveraging this tactic was proposed by \cite{meng2017deep}, employing a generative model for keyword prediction with a recurrent encoder-decoder framework with an attention mechanism capable of detecting keywords in the input text sequence and also potentially finding keywords that do not appear in the text. Since finding absent keywords involves a very hard problem of finding a correct class in a set of usually thousands of unbalanced classes, their model also employs a copying mechanism \citep{gu2016incorporating} based on positional information, in order to allow the model to find important keywords present in the text, which is a much easier problem.

Very recently, the model proposed by \cite{meng2017deep} has been somewhat improved by investigating different ways in which the target keywords can be fed to a classifier during the training phase. While the original system used a so-called \textit{one-to-one} approach, where a training example consists of an input text and a single keyword, the improved model \citep{meng2019does} now employs a \textit{one-to-seq} approach, where an input text is matched with a concatenated sequence made of all the keywords for a specific text. The study also shows that the order of the keywords in the text matters. The best performing model from \cite{meng2019does}, named CopyRNN, is used in our experiments for the comparison with the state-of-the-art (see Section \ref{sec:experimental-setting}). A \textit{one-to-seq} approach has been even further improved by \cite{yuan2019one}, who incorporated two diversity mechanisms into the model. The mechanisms (called \textit{semantic coverage} and \textit{orthogonal regularization})  constrain the over-all inner representation of a generated keyword sequence to be semantically similar to the overall meaning of the source text and therefore force the model to produce diverse keywords. The resulting model leveraging these mechanisms has been named CatSeqD and is also used in our experiments for the comparison between TNT-KID and the state-of-the-art.

A further improvement of the generative approach towards keyword detection has been proposed by \citet{chan2019neural}, who integrated a reinforcement learning (RL) objective into the keyphrase  generation approach proposed by \cite{yuan2019one}. This is done by introducing an adaptive reward function that encourages the model to generate sufficient amount of accurate keyphrases. They also propose a new Wikipedia based evaluation method that can more robustly evaluate the quality of the predicted keyphrases by also considering name variations of the ground-truth keyphrases.

We are aware of one study that tackled keyword detection with transformers. \citet{sahrawat2019keyphrase} fed contextual embeddings generated using several transformer and recurrent architectures (BERT \citep{devlin2018bert}, RoBERTa \citep{liu2019roberta}, GPT-2 \citep{radford2019gpt2}, ELMo \citep{peters2018deep}, etc.) into two distinct neural architectures, a bidirectional Long short-term memory network (BiLSTM) and a BiLSTM network with an additional Conditional random fields layer (BiLSTM-CRF). Same as in \cite{gollapalli2017incorporating}, they formulate a keyword extraction task as a sequence labelling approach, in which each word in the document is assigned one of the three possible labels: $k_b$ denotes that the word is the first word in a keyphrase, $k_i$ means that the word is inside a keyphrase, and $k_o$ indicates that the word is not part of a keyphrase.

The study shows that contextual embeddings generated by transformer architectures generally perform better than static (e.g., FastText embeddings \citep{bojanowski2017enriching}) and among them BERT showcases the best performance. Since all of the keyword detection experiments are conducted on scientific articles, they also test SciBERT \citep{beltagy2019scibert}, a version of BERT pretrained on a large multi-domain corpus of scientific publications containing 1.14M papers sampled from Semantic Scholar. They observe that this genre specific pretraining on texts of the same genre as the texts in the keyword datasets, slightly improves the performance of the model. They also report significant gains in performance when the BiLSTM-CRF architecture is used instead of BiLSTM.

The neural sequence-to-sequence models are capable of outperforming all older supervised and unsupervised models by a large margin but do require a very large training corpora with tens of thousands of documents for successful training. This means that their use is limited only to languages (and genres) in which large corpora with manually labeled keywords exist. On the other hand, the study by \citet{sahrawat2019keyphrase} indicates that the employment of contextual embeddings reduces the need for a large dataset with manually labeled keywords. These models can therefore be deployed directly on smaller datasets by leveraging semantic information already encoded in contextual embeddings.

\subsection{Unsupervised keyword extraction methods}
\label{Sec:UnupervisedSOTA}

The previous section discussed recently emerged methods for keyword extraction that operate in a supervised learning setting and can be data-intensive and time consuming. Unsupervised keyword detectors can tackle these two problems, yet at the cost of the reduced overall performance.

Unsupervised approaches need no training and can be applied directly without relying on a gold standard document collection. They can be divided into statistical and graph-based methods:

\begin{itemize}
\item Statistical methods, such as KP-MINER \citep{el2009kp}, RAKE \citep{rose2010automatic} and YAKE \citep{campos2,campos2018yake}, use statistical characteristics of the texts to capture keywords.
\item Graph-based methods, such as TextRank \citep{mihalcea2004textrank}, Single Rank \citep{wan2008single}, TopicRank \citep{bougouin2013topicrank}, Topical PageRank \citep{sterckx2015topical} and RaKUn \citep{vskrlj2019rakun} build graphs to rank words based on their position in the graph.
\end{itemize}

Among the statistical approaches, the state-of-the-art keyword extraction algorithm is YAKE \citep{campos2,campos2018yake}. It defines a set of features capturing keyword characteristics which are heuristically combined to assign a single score to every keyword. These features include casing, position, frequency, relatedness to context and dispersion of a specific term.

One of the first graph-based methods for keyword detection is TextRank \citep{mihalcea2004textrank}, which first extracts a lexical graph from text documents and then leverages Google's PageRank algorithm to rank vertices in the graph according to their importance inside a graph. This approach was somewhat upgraded by TopicRank \citep{bougouin2013topicrank}, where candidate keywords are additionally clustered into topics and used as vertices in the graph. Keywords are detected by selecting a candidate from each of the top-ranked topics. The most recent graph-based keyword detector is RaKUn \citep{vskrlj2019rakun} that employs several new techniques for graph construction and vertice ranking. First, initial lexical graph is expanded and adapted with the introduction of meta-vertices, i.e., aggregates of existing vertices. Second, for keyword detection and ranking, a graph-theoretic \textit{load centrality} measure is used along with the implemented graph redundancy filters.

\section{Methodology}
\label{sec:methodology}

This section presents the methodology of our approach. Section \ref{sec:architecture} presents the architecture of the neural model, Section \ref{sec:transfer} covers the transfer learning techniques used, Section \ref{sec:kw-identification} explains how the final fine-tuning phase of the keyword detection workflow is conducted and Section \ref{sec:eval} covers evaluation of the model.

\subsection{Architecture}
\label{sec:architecture}

\begin{figure}
    \centering
    \begin{tabular}{cc}
    \subcaptionbox{Model architecture. \label{fig:architecture}}{\includegraphics[width = 1.8in]{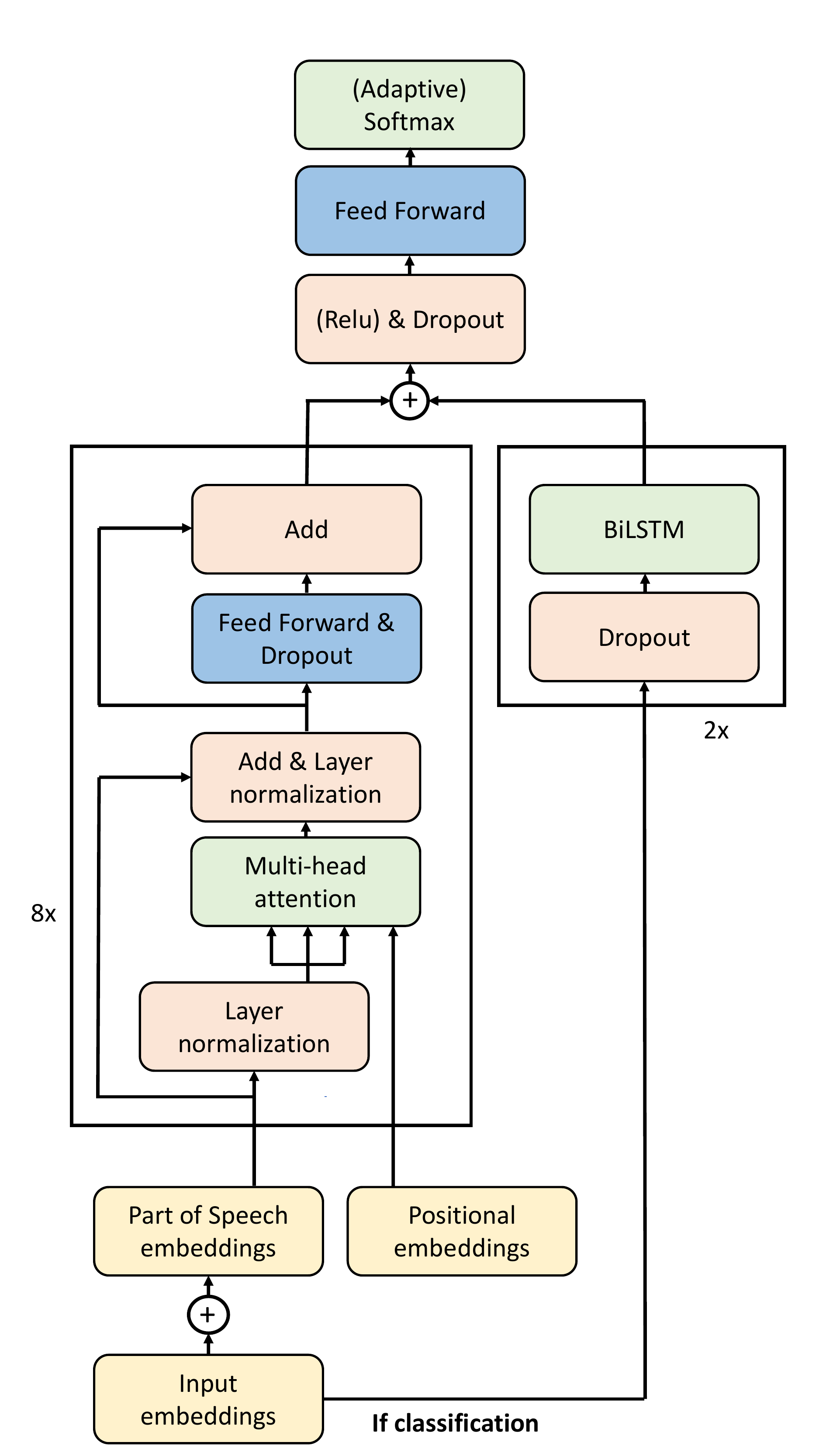}} &
    \subcaptionbox{The attention mechanism. \label{fig:attention}}{\includegraphics[width = 1.9in]{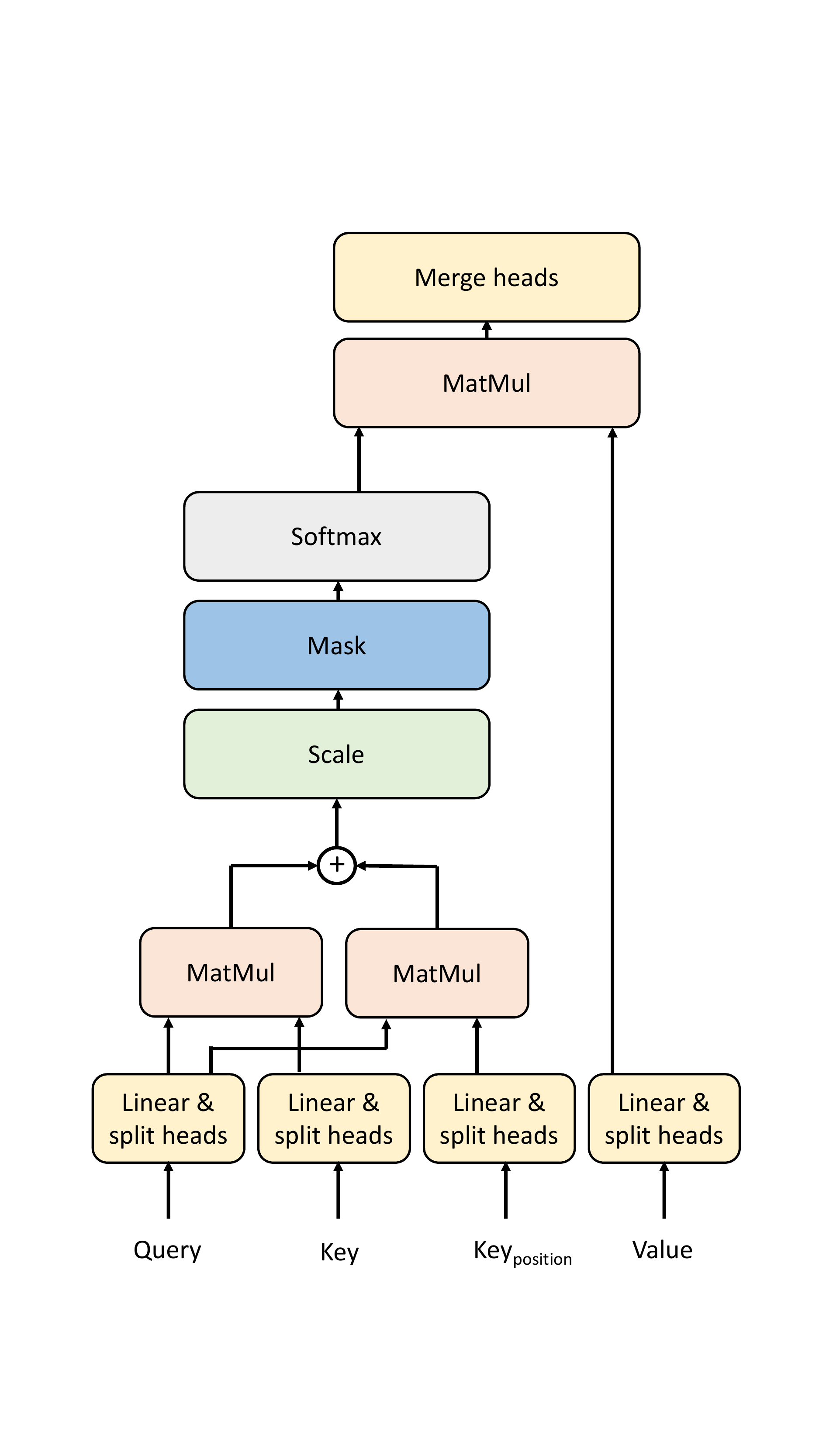}}
    \end{tabular}
    \caption{TNT-KID's architecture overview.}
\end{figure}

The model follows an architectural design of an original transformer encoder \citep{vaswani2017attention} and is presented in Figure \ref{fig:architecture}. Same as in the GPT-2 architecture \citep{radford2019gpt2}, the encoder consists of a normalization layer that is followed by a multi-head attention mechanism. A residual connection is employed around the attention mechanism, which is followed by another layer normalization. This is followed by the fully connected feed-forward and dropout layers, around which another residual connection is employed.

For two distinct training phases, language model pretraining and fine-tuning, two distinct ``heads'' are added on top of the encoder, which is identical for both phases and therefore allows for the transfer of weights from the pretraining phase to the fine-tuning phase. The language model head predicts the probability for each word in the vocabulary that it appears at a specific position in the sequence and consists of a dropout layer and a feed forward layer of size $\textrm{SL} * |V|$, where $\textrm{SL}$ stands for sequence length (i.e., a number of words in the input text) and $|V|$ stands for the vocabulary size. This is followed by the adaptive softmax layer \citep{grave2017efficient} (see description below). 

During fine-tuning, the language model head is replaced with a token classification head, in which we apply ReLu non-linearity and dropout to the encoder output, and then feed the output to the feed forward classification layer of size $\textrm{SL} * \textrm{NC}$, where NC stands for the number of classes (in our case 2, since we model keyword extraction as a binary classification task, see Section 3.3 for more details). Finally, a softmax layer is added in order to obtain probabilities for each class.

We also propose some significant modifications of the original GPT-2 architecture. First, we propose a re-parametrization of the attention mechanism that allows to model the relation between a token and its position (see Figure \ref{fig:attention}). Note that standard scaled dot-product attention \citep{vaswani2017attention} requires three inputs, a so-called \textit{query, key, value} matrix representations of the embedded input sequence and its positional information (i.e., element wise addition of input embeddings and positional embeddings) and the idea is to obtain attention scores (in a shape of an attention matrix) for each relation between tokens inside these inputs by first multiplying \textit{query} ($Q$) and transposed \textit{key} ($K$) matrix representations, applying scaling and softmax functions, and finally multiplying the resulting normalized matrix with the \textit{value} ($V$) matrix, or more formally,

\[ \textrm{Attention}(Q,K,V) = \textrm{softmax} \bigg ( \frac{QK^T}{\sqrt{d_k}} \bigg ) V \]

where $d_k$ represents the scaling factor, usually corresponding to the first dimension of the \textit{key} matrix. 
On the other hand, we propose to add an additional positional input representation matrix $K_{\textrm{position}}$ and model attention with the following equation:

\[ \textrm{Attention}(Q,K,V, K_{pos}) = \textrm{softmax} \bigg ( \frac{QK^T + QK_{position}^T}{\sqrt{d_k}} \bigg ) V \]

The reason behind this modification is connected with the hypothesis, that token position is important in the keyword identification task and with this re-parametrization the model is capable of directly modelling the importance of relation between each token and each position. Note that we use relative positional embeddings for representing the positional information, same as in \cite{dai2019transformer}, where the main idea is to only encode the relative positional information in the hidden states instead of the absolute. 

Second, besides the text input, we also experiment with the additional part-of-speech (POS) tag sequence as an input. This sequence is first embedded and then added to the word embedding matrix. Note that this additional input is optional and is not included in the model for which the results are presented in Section \ref{sec-KEresults} due to marginal effect on the performance of the model in the proposed experimental setting (see Section \ref{sec:ablation}).

While the modifications presented above affect both training phases (i.e., the language model pretraining and the token classification fine-tuning), the third modification only affects the language model pretraining (see Section \ref{sec:transfer}) and involves replacing the standard input embedding layer and softmax function with adaptive input representations \citep{baevski2018adaptive} and an adaptive softmax \citep{grave2017efficient}. The main idea is to exploit the unbalanced word distribution to form word clusters containing words with similar appearance probabilities. The entire vocabulary is split into a smaller cluster containing about 10 percent of words that appear most frequently, a second slightly bigger cluster that contains words that appear less frequently and a third cluster that contains all the other words that appear rarely in the corpus. During language model training, instead of predicting an entire vocabulary distribution at each time step, the model first tries to predict a cluster in which a target word appears in and after that predicts a vocabulary distribution just for the words in that cluster. Since in a large majority of cases the target word belongs to the smallest cluster containing most frequent words, the model in most cases only needs to generate probability distribution for a tenth of a vocabulary, which drastically reduces the memory requirements and time complexity of the model at the expense of a marginal drop in performance.

We also present the modification, which only affects the fine-tuning token classification phase (see Section \ref{sec:kw-identification}). During this phase, a two layer randomly initialised encoder, consisting of dropout and two bidirectional Long short-term memory (BiLSTM) layers, is added (with element-wise summation) to the output of the transformer encoder. The initial motivation behind this adaptation is connected with findings from the related work which suggest that recurrent layers are quite successful at modelling positional importance of tokens in the keyword detection task \citep{meng2017deep,yuan2019one} and by the study of \cite{sahrawat2019keyphrase}, who also reported good results when a BiLSTM classifier and contextual embeddings generated by transformer architectures were employed for keyword detection. Also, the results of the initial experiments suggested that some performance gains can in fact be achieved by employing this modification.

In terms of computational complexity, a self-attention layer complexity is $O(n^2 * d)$ and the complexity of the recurrent layer is $O(n * d^2)$, where $n$ is the sequence length and d is the embedding size \citep{vaswani2017attention}. The standard TNT-KID model employs sequence size of 256, embedding size of 512 and 8 attention layers. The complexity of the model without recurrent encoder is therefore $256^2 * 512 * 8 = 268435456$. By adding the recurrent encoder with two recurrent bidirectional layers (which is the same as adding 4 recurrent layers, since each bidirectional layer contains two unidirectional LSTM layers), the complexity increases by $256 * 512^2 * 4 = 268435456$, which means that the model with the additional recurrent encoder conducts token classification roughly two times slower than the model without the encoder. Note that this addition does not affect the language model pretraining, which tends to be the more time demanding task due to larger corpora involved.

Finally, we also experiment with an employment of the BiLSTM-CRF classification head on top of the transformer encoder, in order to compare our proposed approach to the approach proposed by \cite{sahrawat2019keyphrase} (see Section \ref{sec:ablation} for more details about the results of this experiment). For this experiment, during the fine-tuning token classification phase, the token classification head described above is replaced with a BiLSTM-CRF classification head proposed by \cite{sahrawat2019keyphrase}, containing one BiLSTM layer and a CRF \citep{lafferty2001conditional} layer.\footnote{Note that in the experiments in which we employ BiLSTM-CRF, we do not add an additional two layer BiLSTM encoder described above to the output of the transformer encoder.} Outputs of the BiLSTM $f={f_1,...,f_n}$ are fed as inputs to a CRF layer, which returns the output score $s(f,y)$ for each possible label sequence according to the following equation:

\[s(f,y) = \sum_{t=1}^{n} \tau_{y_{t-1},y_t} + f_{t,y_t} \]

$\tau_{y_{t-1},y_t}$ is a transition matrix representing the transition score from class $y_{t-1}$ to $y_t$. The final probability of each label sequence score is generated by exponentiating the scores and normalizing over all possible output label sequences:

\[p(y|f)=\frac{exp(s(f,y))}{\sum_{y'} exp(s(f',y'))}  \]

To find the optimal sequence of labels efficiently, the CRF layer uses the Viterbi algorithm \citep{forney1973viterbi}.

\subsection{Transfer learning}
\label{sec:transfer}

Our approach relies on a transfer learning technique \citep{howard2018universal, devlin2018bert}, where a neural model is first pretrained as a language model on a large corpus. This model is then fine-tuned for each specific keyword detection task on each specific manually labeled corpus by adding and training the token classification head described in the previous section. With this approach, the syntactic and semantic knowledge of the pretrained language model is transferred and leveraged in the keyword detection task, improving the detection on datasets that are too small for the successful semantic and syntactic generalization of the neural model.

In the transfer learning scenario, two distinct pretraining objectives can be considered. First is the autoregressive language modelling where the task can be formally defined as predicting a probability distribution of words from the fixed size vocabulary $V$, for word $w\textsubscript{t}$, given the historical sequence $w\textsubscript{1:t-1} = [w_1,...,w_{t-1}]$. This pretraining regime was used in the GPT-2 model \citep{radford2019gpt2} that we modified. Since in the standard transformer architecture self-attention is applied to an entire surrounding context of a specific word (i.e., the words that appear after a specific word in each input sequence are also used in the self-attention calculation), we employ obfuscation masking to the right context of each word when the autoregressive language model objective is used, in order to restrict the information only to the prior words in the sentence (plus the word itself) and prevent target leakage (see \citet{radford2019gpt2} for details on the masking procedure). 

Another option is a masked language modelling objective, first proposed by \cite{devlin2018bert}. Here, a percentage of words from the input sequence is masked in advance, and the objective is to predict these masked words from an unmasked context. This allows the model to leverage both left and right context, or more formally, the token $w\textsubscript{t}$ is also determined by sequence of tokens $w\textsubscript{t+1:n} = [w_{t+1},...,w_{t+n}]$. We follow the masking procedure described in the original paper by \cite{devlin2018bert}, where 15 percent of words are randomly designated as targets for prediction, out of which 80 percent are replaced by a masked token ($<mask>$), 10 percent are replaced by a random word and 10 percent remain intact.

The final output of the model is a softmax probability distribution calculated over the entire vocabulary, containing the predicted probabilities of appearance (P) for each word given its left (and in case of the \textit{masked language modelling objective} also right) context. Training therefore consists of the minimization of the negative log-loss (NLL) on the batches of training corpus word sequences by backpropagation through time:

\begin{equation}
\textrm{NLL} = -\sum_{i=1}^{n}\log{P(w_i|w\textsubscript{1:i-1})}
\label{eq:NLL}
\end{equation}

While the \textit{masked language modelling} objective might outperform autoregressive language modelling objective in a setting where a large pretraining corpus is available \citep{devlin2018bert} due to the inclusion of the right context, these two training objectives have at least to our knowledge never been compared in a setting where only a relatively small domain specific corpus is available for the pretraining phase. For more details about the performance comparison of these two pretraining objectives, see Section \ref{sec:ablation}.

\subsection{Keyword identification}
\label{sec:kw-identification}

\begin{figure}[t!]
\vspace{1.0cm}
\centering
  \includegraphics[width=\linewidth]{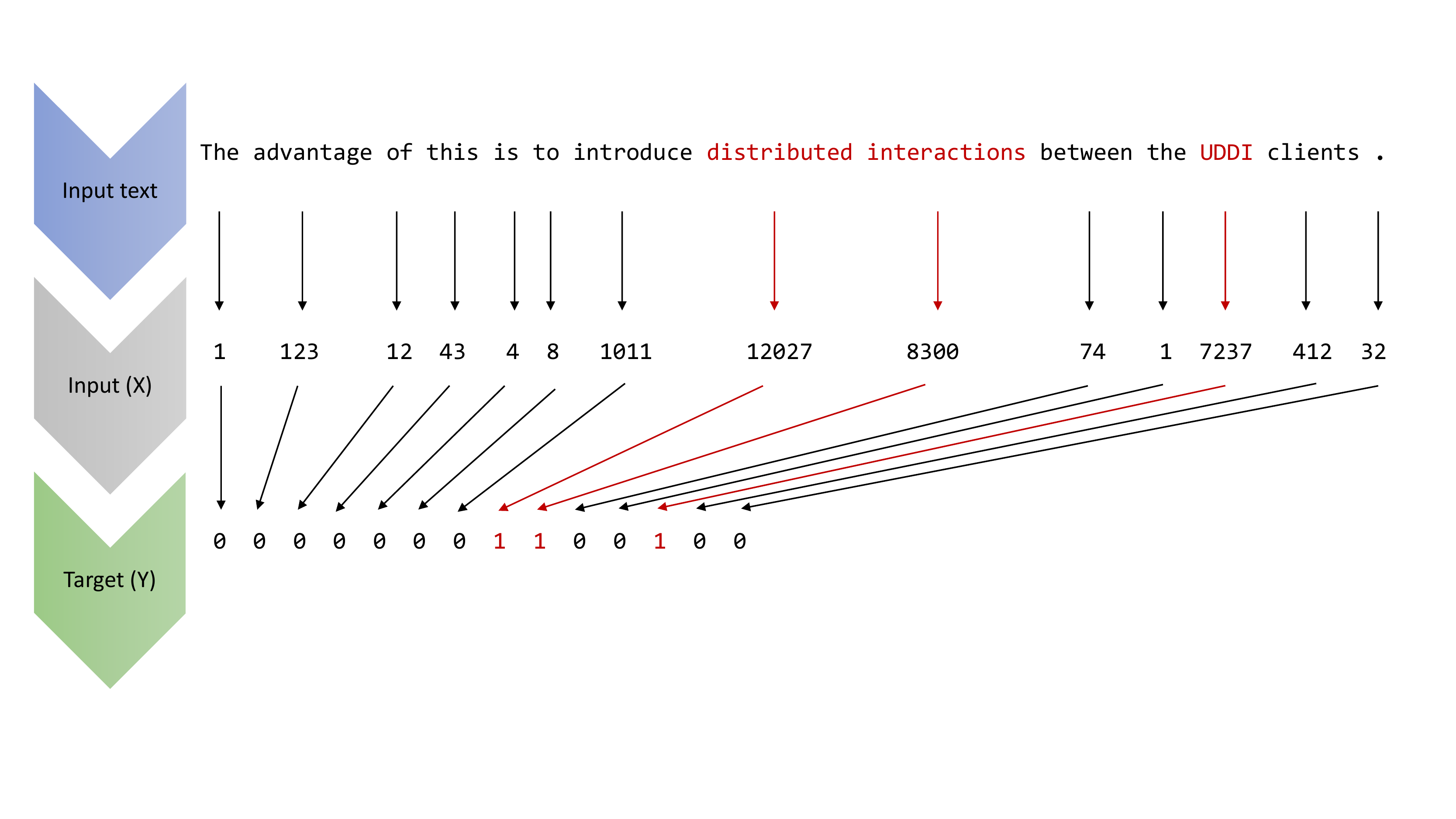}
  \caption{Encoding of the input text ``\textit{The advantage of this is to introduce distributed interactions between the UDDI clients.}'' with keywords \textit{distributed interactions} and \textit{UDDI}. In the first step, the text is converted into a numerical sequence, which is used as an input to the model. The model is trained to convert this numerical sequence into a sequence of zeroes and ones, where the ones indicate the position of a keyword.}
  \label{input-example}

\end{figure}

Since each word in the sequence can either be a keyword (or at least part of the keyphrase) or not, the keyword tagging task can be modeled as a binary classification task, where the model is trained to predict if a word in the sequence is a keyword or not.\footnote{Note that this differs from the sequence labelling approach proposed by \cite{sahrawat2019keyphrase}, where each word in the document is assigned one of three possible labels (see Section \ref{sec-relatedwork} for details).} Figure \ref{input-example} shows an example of how an input text is first transformed into a numerical sequence that is used as an input of the model, which is then trained to produce a sequence of zeroes and ones, where the positions of ones indicate the positions of keywords in the input text. 

Since a large majority of words in the sequence are not keywords, the usage of a standard NLL function (see equation \ref{eq:NLL}), which would simply calculate a sum of log probabilities that a word is either a keyword or not for every input word sequence, would badly affect the recall of the model since the majority negative class would prevail. To solve this problem and maximize the recall of the system, we propose a custom classification loss function, where probabilities for each word in the sequence are first aggregated into two distinct sets, one for each class. For example, text ``\textit{The advantage of this is to include distributed interactions between the UDDI clients.}'' in Figure \ref{input-example} would be split into two sets, first one containing probabilities for all the words in the input example which are not keywords (\textit{The, advantage, of, this, is, to, include, between, the, clients, .}), and the other containing probabilities for all the words in the input example that are keywords or part of keyphrases (\textit{distributed, interactions, UDDI}). Two NLLs are calculated, one for each probability set, and both are normalized with the size of the set. Finally, the NLLs are summed. More formally, the loss is computed as follows. Let $W = \{w_i\}_{i = 1}^{n}$ represent an enumerated sequence of tokens for which predictions are obtained. Let $p_i$ represent the predicted probabilities for the $i$-th token that it either belongs or does not belong to the ground truth class. The $o_i$ represents the output weight vector of the neural network for token $i$ and $j$ corresponds to the number of classes (two in our case as the word can be a keyword or not). Predictions are in this work obtained via a log-softmax transform ($lso$), defined as follows (for the $i$-th token):

\begin{equation*}
    p_i = \textrm{lso}(o_i) = \log \frac{\exp(o_i)}{\sum_j \exp(o_j)}.
\end{equation*}

 The loss function is comprised from two main parts. Let $K_+ \subseteq W$ represent tokens that are keywords and $K_- \subseteq W$ the set of tokens that are \textbf{not} keywords. Note that $|K_- \cup K_+| = n$, i.e., the two sets cover all considered tokens for which predictions are obtained.
 During loss computation, only the probabilities of the ground truth class are considered. We mark them with $p_i^+$ or $p_i^-$.
 Then the loss is computed as
\begin{equation*}
    L_+ = -\frac{1}{|K_+|}\sum_{w_i \in K_+} p_i^+ \quad  \textrm{and} \quad L_- = - \frac{1}{|K_-|}\sum_{w_i \in K_-} p_i^-.
\end{equation*}
The final loss is finally computed as:
\begin{equation*}
    \textrm{Loss} = L_+ + L_-.
\end{equation*}

Note that even though all predictions are given as an argument, the two parts of the loss address different token indices ($i$).

In order to produce final set of keywords for each document, tagged words are extracted from the text and duplicates are removed. Note that a sequence of ones is always interpreted as a multi-word keyphrase and not as a combination of one-worded keywords (e.g., \textit{distributed interactions} from Figure \ref{input-example} is considered as a single multi-word keyphrase and not as two distinct one word keywords). After that, the following filtering is conducted:

\begin{itemize}
    \item If a keyphrase is longer than four words, it is discarded.
    \item Keywords containing punctuation (with the exception of dashes and  apostrophes) are removed.
    \item The detected keyphrases are ranked and arranged according to the softmax probability assigned by the model in a descending order.
\end{itemize}

\subsection{Evaluation}
\label{sec:eval}

To asses the performance of the model, we measure F1@$k$ score, a harmonic mean between Precision@$k$ and Recall@$k$.

In a ranking task, we are interested in precision at rank $k$. This means that only the keywords ranked equal to or better than $k$ are considered and the rest are disregarded. Precision is the ratio of the number of correct keywords returned by the system divided by the number of all keywords returned by the system, or more formally:

\[precision = \frac{|correct~~ returned~~ keywords@k|}{|returned~~ keywords|}\]

Recall@$k$ is the ratio of the number of correct keywords returned by the system and ranked equal to or better than $k$ divided by the number of correct ground truth keywords:
    
 \[recall = \frac{|correct~~ returned~~ keywords@k|}{|correct~~ keywords|}\]  

Due to the high variance of a number of ground truth keywords, this type of recall becomes problematic if $k$ is smaller than the number of ground truth keywords, since it becomes impossible for the system to achieve a perfect recall. (Similar can happen to precision@k, if the number of keywords in a gold standard is lower than \textit{k}, and returned number of keywords is fixed at \textit{k}.) We shall discuss how this affects different keyword detection systems in Section \ref{sec:conclusion}.
    
Finally, we formally define F1@$k$ as a harmonic mean between Precision@$k$ and Recall@$k$:

\[ F1@k= 2 * \frac{P@k * R@k} {P@k + R@k} \]

In order to compare the results of our approach to other state-of-the-art approaches, we use the same evaluation methodology as \cite{yuan2019one} and \cite{meng2019does}, and measure F1@$k$ with $k$ being either 5 or 10. Note that F1@$k$ is calculated as a harmonic mean of macro-averaged precision and recall, meaning that precision and recall scores for each document are averaged and the F1 score is calculated from these averages. Same as in the related work, lowercasing and stemming are performed on both the gold standard and the generated keywords (keyphrases) during the evaluation. Only keywords that appear in the text of the documents (present keywords)\footnote{Note that scientific and news articles often list keywords that do not appear in the text of the article. For example, an NLP paper would often list ``\textit{Text mining}'' as a keyword of the paper, even though the actual phrase does not appear in the text of the paper.} were used as a gold standard and the documents containing no present keywords were removed, in order to make the results of the conducted experiments comparable with the reported results from the related work.

\section{Experiments}
\label{sec:experimental-setting}

We first present the datasets used in the experiments. This is followed by the experimental design and the results achieved by TNT-KID in comparison to the state-of-the-art.

\subsection{Keyword extraction datasets}
\label{sec:datasets}

Experiments were conducted on seven datasets from two distinct genres, scientific papers about computer science and news. The following datasets from the computer science domain are used:

\begin{itemize}
    \item \textbf{KP20k \citep{meng2017deep}}:  This dataset contains titles, abstracts, and keyphrases of 570,000 scientific articles from the field of computer science. The dataset is split into train set (530,000), validation set (20,000) and test set (20,000).
    
    \item \textbf{Inspec \citep{hulth2003improved}}: The dataset contains 2,000 abstracts of scientific journal papers in computer science collected between 1998 and 2002. Two sets of keywords are assigned to each document, the controlled keywords that appear in the Inspec thesaurus, and the uncontrolled keywords, which are assigned by the editors. Only uncontrolled keywords are used in the evaluation, same as by \cite{meng2017deep}, and the dataset is split into 500 test papers and 1500 train papers. 
    
    \item \textbf{Krapivin \citep{krapivin2009large}}: This dataset contains 2,304 full scientific papers from computer science domain published by ACM between 2003 and 2005 with author-assigned keyphrases. 460 papers from the dataset are used as a test set and the others are used for training.  Only titles and abstracts are used in our experiments.
    
    \item \textbf{NUS \citep{nguyen2007keyphrase}}: The dataset contains titles and abstracts of 211 scientific conference papers from the computer science domain and contains a set of keywords assigned by student volunters and a set of author assigned keywords, which are both used in evaluation.
    \item \textbf{SemEval \citep{Kim:2010:STA:1859664.1859668}}: The dataset used in the SemEval-2010 Task 5, Automatic Keyphrase Extraction from Scientific Articles, contains 244 articles from the computer science domain collected from the ACM Digital Library. 100 articles are used for testing and the rest are used for training. Again, only titles and abstracts are used in our experiments, the article's content was discarded.
    
\end{itemize}

From the news domain, three datasets with manually labeled gold standard keywords are used:

\begin{itemize}
    
     \item \textbf{KPTimes \citep{gallina2019kptimes}}: The corpus contains 279,923 news articles containing editor assigned keywords that were collected by crawling New York Times news website\footnote{\url{https://www.nytimes.com}}. After that, the dataset was randomly divided into training (92.8 percent), development (3.6 percent) and test (3.6 percent) sets.
     
     \item \textbf{JPTimes \citep{gallina2019kptimes}}: Similar as \textbf{KPTimes}, the corpus was collected by crawling Japan Times online news portal\footnote{\url{https://www.japantimes.co.jp}}. The corpus only contains 10,000 English news articles and is used in our experiments as a test set for the classifiers trained on the \textbf{KPTimes} dataset.
    
    \item \textbf{DUC \citep{wan2008single}}: The dataset consists of 308 English news articles and contains 2,488 hand labeled keyphrases.
    
\end{itemize}

The statistics about the datasets that are used for training and testing of our models are presented in Table \ref{tbl:datasets}. Note that there is a big variation in dataset sizes in terms of number of documents (column \textit{No. docs}), and in an average number of keywords (column \textit{Avg. kw.}) and present keywords per document (columns \textit{Avg. present kw.}), ranging from 2.35 present keywords per document in \textit{KPTimes-valid} to 7.79 in \textit{DUC-test}.

\begin{table}[t]
\centering
\caption{Datasets used for empirical evaluation of keyword extraction algorithms. \textit{No.docs} stands for number of documents, \textit{Avg. doc. length} stands for average document length in the corpus, \textit{Avg. kw.} stands for average number of keywords per document in the corpus, \textit{\% present kw.} stands for the percentage of keywords that appear in the corpus (i.e., percentage of document's keywords that appear in the text of the document) and \textit{Avg. present kw.} stands for the average number of keywords per document that actually appear in the text of the specific document.}
\resizebox{0.99\textwidth}{!}{
\begin{tabular}{llllll}
\hline
Dataset & No. docs & Avg. doc. length & Avg. kw. & \% present kw. & Avg. present kw.\\
\hline
\textbf{Computer science papers} & & & & & \\\hline 
KP20k-train & 530,000 & 156.34 & 5.27 & 62.43 & 3.29 \\
KP20k-valid & 20,000 & 156.55 & 5.26 & 62.30 & 3.28 \\
KP20k-test & 20,000 & 156.52 & 5.26 & 62.55 & 3.29 \\
Inspec-valid & 1500 & 125.21 & 9.57 & 76.92 & 7.36 \\
Inspec-test & 500 & 121.82 & 9.83 & 78.14 & 7.68 \\
Krapivin-valid & 1844 & 156.65 & 5.24 & 54.34 & 2.85 \\
Krapivin-test & 460 & 157.76 & 5.74 & 55.66 & 3.20 \\
NUS-test & 211 & 164.80 & 11.66 & 50.47 & 5.89 \\
SemEval-valid & 144 & 166.86 & 15.67 & 45.43 & 7.12 \\
SemEval-test & 100 & 183.71 & 15.07 & 44.53 & 6.71 \\\hline
\textbf{News articles} & & & & & \\\hline 
KPTimes-train & 259,923 & 783.32 & 5.03 & 47.30 & 2.38 \\
KPTimes-valid & 10,000 & 784.65 & 5.02 & 46.78 & 2.35 \\
KPTimes-test & 10,000 & 783.47 & 5.04 & 47.59 & 2.40 \\
JPTimes-test & 10,000 & 503.00 & 5.03 & 76.73 & 3.86 \\
DUC-test & 308 & 683.14 & 8.06 & 96.62 & 7.79 \\
\hline
\end{tabular}
}
\label{tbl:datasets}
\end{table}

\subsection{Experimental design}

We conducted experiments on the datasets described in Section \ref{sec:datasets}. First, we lowercased and tokenized all datasets. We experimented with two tokenization schemes, word tokenization and Sentencepiece \citep{kudo2018sentencepiece} byte-pair encoding (see Section \ref{sec:ablation} for more details on how these two tokenization schemes affect the overall performance). During both tokenization schemes, a special $<eos>$ token is used to indicate the end of each sentence. For the best performing model, for which the results are presented in Section \ref{sec-KEresults}, byte-pair encoding was used. For generating the additional POS tag sequence input described in Section \ref{sec:architecture}, which was \textbf{not} used in the best performing model, Averaged Perceptron Tagger from the NLTK library \citep{loper2002nltk} was used. The neural architecture was implemented in PyTorch \citep{paszke2019pytorch}.  

In the pretraining phase, two language models were trained for up to ten epochs, one on the concatenation of all the texts from the computer science domain and the other on the concatenation of all the texts from the news domain. Overall the language model train set for computer science domain contained around 87 million tokens and the news train set about 232 million tokens. These small sizes of the language model train sets enable relatively fast training and smaller model sizes (in terms of number of parameters) due to the reduced vocabulary.

After the pretraining phase, the trained language models were fine-tuned on each dataset's \textit{validation} sets (see Table \ref{tbl:datasets}), which were randomly split into 80 percent of documents used for training and 20 percent of documents used for validation. The documents containing more than 256 tokens are truncated, while the documents containing less than 256 tokens are padded with a special $<pad>$ token at the end. Each model was fine-tuned for a maximum of 10 epochs and after each epoch the trained model was tested on the documents chosen for validation. The model that showed the best performance on this set of validation documents (in terms of F@10 score) was used for keyword detection on the test set. Validation sets were also used to determine the best hyperparameters of the model and all combinations of the following hyperparameter values were tested before choosing the best combination, which is written in bold in the list below and on average worked best for all the datasets in both domains\footnote{Note that the same set of hyperparameters are also used in the pretraining phase.}:

\begin{itemize}

\item Learning rates: 0.00005, 0.0001, \textbf{0.0003}, 0.0005, 0.001
\item Embedding size: 256, \textbf{512}
\item Number of attention heads: 4, \textbf{8}, 12
\item Sequence length: 128, \textbf{256}
\item Number of attention layers: 4, \textbf{8}, 12 
\end{itemize}

Note that in our experiments, we use the same splits as in related work \citep{meng2019does,meng2017deep,gallina2019kptimes} for all datasets with predefined splits (see Table \ref{tbl:datasets}). The exceptions are NUS, DUC and JPTimes datasets with no available predefined validation-test splits. For NUS and DUC, 10-fold cross-validation is used and the model used for keyword detection on the JPTimes-test dataset was fine-tuned on the KPTimes-valid dataset. Another thing to consider is that in the related work by \cite{yuan2019one}, \cite{meng2017deep} and \cite{gallina2019kptimes}, to which we are comparing, large datasets KPTimes-train and KP20k-train with 530,000 documents and 260,00 documents, respectively, are used for the classification model training and these trained models are applied on all test sets from the matching domain. On the other hand, we do not train our classification models on these two large train sets but instead use smaller KPTimes-valid and KP20k-valid datasets for training, since we argue that, due to language model pretraining, fine-tuning the model on a relatively small labeled dataset is sufficient for the model to achieve competitive performance. We do however conduct the language model pretraining on the concatenation of all the texts from the computer science domain and the news domain as explained above, and these two corpora also contain texts from KPTimes-train and KP20k-train datasets.

\subsection{Keyword extraction results and comparison to the state-of-the-art}
\label{sec-KEresults}

In Table \ref{tbl:results}, we present the results achieved by TNT-KID and a number of algorithms from the related work on the datasets presented in Table \ref{tbl:datasets}. Evaluation measures were presented in Section \ref{sec:eval}. Only keywords which appear in a text (present keywords) were used as a gold standard in order to make the results of the conducted experiments comparable with reported results from the related work. Note that TfIdf, TextRank, YAKE and RaKUn algorithms are unsupervised and do not require any training, KEA, Maui, GPT-2, GPT-2 + BiLSTM-CRF and TNT-KID were trained on the different \textit{validation} set for each of the datasets, and CopyRNN and CatSeqD were trained on the large KP20k-train dataset for keyword detection on computer science domain, and on the KPTimes-train dataset for keyword detection on the news domain, since they require a large train set for competitive performance.  

For RaKUn \citep{vskrlj2019rakun} and YAKE \citep{campos2020yake} we report results for default hyperparameter settings, since the authors of RaKUn, as well as YAKE's authors claim that a single hyperparameter set can offer sufficient performance across multiple datasets. We used the author's official github implementations\footnote{\url{https://github.com/SkBlaz/rakun} and \url{https://github.com/LIAAD/yake}} in the experiments. For KEA and Maui we do not conduct additional testing on corpora for which results are not available in the related work (KPTimes, JPTimes and DUC corpus) due to bad performance of the algorithms on all the corpora for which results are available. Finally, for TfIdf and TextRank we report results from the related work where available \citep{yuan2019one} and use the implementation of the algorithms from the Python Keyphrase Extraction (PKE) library\footnote{\url{https://github.com/boudinfl/pke}} to generate unavailable results. Same as for RaKUn and YAKE, default hyperparameters are used.

For KEA, Maui, CopyRNN and CatSeqD, we report results for the computer science domain published in \cite{yuan2019one} and for the news domain we report results for CopyRNN published in \cite{gallina2019kptimes}. The results that were not reported in the related work are results for CatSeqD on KPTimes, JPTimes and DUC, since this model was originally not tested on these three datasets, and the F1@5 score results for CopyRNN on KPTimes and JPTimes. Again, author's official github implementations\footnote{\url{https://github.com/memray/OpenNMT-kpg-release}} were used for training and testing of both models. The models were trained and tested on the large KPTimes-train dataset with a help of a script supplied by the authors of the papers. Same hyperparameters that were used for KP20k training in the original papers \citep{yuan2019one,meng2019does} were used.  

We also report results for the unmodified pretrained GPT-2 \citep{radford2019gpt2} model with a standard feed forward token classification head, and a pretrained GPT-2 with a BiLSTM-CRF token classification head, as proposed in \cite{sahrawat2019keyphrase} and described in Section \ref{sec:architecture}\footnote{We use the implementation of GPT-2 from the Transformers library (\url{https://github.com/huggingface/transformers}) and use the Pytorch-crf library (\url{https://pytorch-crf.readthedocs.io/en/stable/}) for the implementation of the BiLSTM-CRF token classification head.}. For these two models, we apply the same fine-tuning regime as for TNT-KID, i.e. we fine-tune the models for up to 10 epoch on each dataset's \textit{validation} sets (see Table \ref{tbl:datasets}), which were randomly split into 80 percent of documents used for training and 20 percent of documents used for validation. The model that showed the best performance on this set of validation documents (in terms of F@10 score) was used for keyword detection on the test set. We use the default hyperparameters for both models and the original GPT-2 tokenization regime.

\begin{table}[t!]
    \centering
        \caption{Empirical evaluation of state-of-the-art keyword extractors. Results marked with * were obtained by our implementation or reimplementation of the algorithm and results without * were reported in the related work.}
    \resizebox{0.99\textwidth}{!}{
    \begin{tabular}{l|cccc|ccccccc}
    & \multicolumn{4}{c}{Unsupervised approaches} &  \multicolumn{5}{c}{Supervised approaches}\\\hline
    & TfIdf & TextRank & YAKE & RaKUn & KEA & Maui & CopyRNN & CatSeqD & GPT-2 & \begin{tabular}[x]{@{}c@{}}GPT-2 + \\ BiLSTM-CRF\end{tabular} & TNT-KID \\\hline
    
    & \multicolumn{9}{c}{\textbf{KP20k}} \\\hline
    F1@5 & 0.072 & 0.181 & 0.141* & 0.177* & 0.046 & 0.005 & 0.317 & \textbf{0.348} & 0.252* & 0.339* & 0.342*\\
    F1@10 & 0.094 & 0.151 & 0.146* & 0.160* & 0.044 & 0.005 & 0.273 & 0.298 & 0.256* & 0.342* & \textbf{0.346*}\\
    \hline
    & \multicolumn{9}{c}{\textbf{Inspec}} \\\hline
    F1@5 & 0.160 & 0.286 & 0.204* & 0.101* & 0.022 & 0.035 & 0.244 & 0.276 & 0.293* & \textbf{0.467*} & 0.447*\\
    F1@10 & 0.244 & 0.339 & 0.223* & 0.108* & 0.022 & 0.046 & 0.289 & 0.333 & 0.335* & \textbf{0.525*} & \textbf{0.525*}\\
    \hline
    & \multicolumn{9}{c}{\textbf{Krapivin}} \\\hline
    F1@5 & 0.067 & 0.185 & 0.215* & 0.127* & 0.018 & 0.005 & 0.305 & \textbf{0.325} & 0.210* & 0.280* & 0.301*\\
    F1@10 & 0.093 & 0.160 & 0.196* & 0.106* & 0.017 & 0.007 & 0.266 & 0.285 & 0.214* & 0.283* & \textbf{0.307*}\\
    \hline
    & \multicolumn{9}{c}{\textbf{NUS}} \\\hline
    F1@5 & 0.112 & 0.230 & 0.159* & 0.224* & 0.073 & 0.004 & 0.376 & \textbf{0.374} & 0.274* & 0.311* & 0.350*\\
    F1@10 & 0.140 & 0.216 & 0.196* & 0.193* & 0.071 & 0.006 & 0.352 & 0.366 & 0.305* & 0.332* & \textbf{0.369*}\\
    \hline
    & \multicolumn{9}{c}{\textbf{SemEval}} \\\hline
    F1@5 & 0.088 & 0.217 & 0.151* & 0.167* & 0.068 & 0.011 & 0.318 & \textbf{0.327} & 0.261* & 0.214 & 0.291*\\
    F1@10 & 0.147 & 0.226 & 0.212* & 0.159* & 0.065 & 0.014 & 0.318 & 0.352 & 0.295* & 0.232 & \textbf{0.355*}\\
    \hline\hline
    & \multicolumn{9}{c}{\textbf{KPTimes}} \\\hline
    F1@5 & 0.179* & 0.022* & 0.105* & 0.168* & * & * & 0.406* & 0.424* & 0.353* & 0.439* & \textbf{0.469*}\\
    F1@10 & 0.151* & 0.030* & 0.118* & 0.139* & * & * & 0.393 & 0.424* & 0.354* & 0.440* & \textbf{0.469*}\\\hline
    & \multicolumn{9}{c}{\textbf{JPTimes}} \\\hline
    F1@5 & 0.266* & 0.012* & 0.109* & 0.225* & * & * & 0.256* & 0.238* & 0.258* & \textbf{0.344*} & 0.337*\\
    F1@10 & 0.229* & 0.026* & 0.135* & 0.185* & * & * & 0.246 & 0.238* & 0.267* & 0.346* & \textbf{0.360*}\\\hline
    & \multicolumn{9}{c}{\textbf{DUC}} \\\hline
    F1@5 & 0.098* & 0.120* & 0.106* & 0.189* & * & * & 0.083 & 0.063* & 0.247* & 0.281* & \textbf{0.312*}\\
    F1@10 & 0.120* & 0.181* & 0.132* & 0.172* & * & * & 0.105 & 0.063* & 0.277* & 0.321* & \textbf{0.355*}\\\hline
    & \multicolumn{9}{c}{\textbf{Average}} \\\hline

    F1@5  & 0.130 & 0.157 & 0.149 & 0.172 & * & * & 0.288 & 0.297 & 0.269* & 0.334* & \textbf{0.356*}\\
    F1@10 & 0.152 & 0.166 & 0.170 & 0.153 & * & * & 0.280 & 0.295 & 0.288* & 0.353* & \textbf{0.386*}\\
    \hline

    \end{tabular}
    }
    \label{tbl:results}
\end{table}

Overall, supervised neural network approaches drastically outperform all other approaches. Among them, TNT-KID performs the best on all eight datasets in terms of F1@10 but is outperformed by CatSeqD (on four datasets) or GPT-2+ BiLSTM-CRF (on two datasets) on six out of eight datasets in terms of F1@5. In terms of F1@10, CatSeqD performs competitively on KP20k, Krapivin, NUS, SemEval and KPTimes datasets but is outperformed by a large margin on three other datasets by both GPT-2 + BiLSTM-CRF and TNT-KID. To be more specific, in terms of F1@10, TNT-KID outperforms the CatSeqD approach by almost 20 percentage points on the Inspec dataset, on the DUC dataset, it outperforms CatSeqD by about 25 percentage points, and on JPTimes it outperforms CatSeqD by about 12 percentage points. 

While the results of CopyRNN are in a large majority of cases very consistent with CatSeqD (CopyRNN performs slightly better than CatSeqD on DUC and JPTimes, and slightly worse on the other six datasets), results of  TNT-KID are very similar to the results of GPT-2 + BiLSTM-CRF. In the majority of cases TNT-KID outperforms GPT-2 + BiLSTM-CRF by a small margin according to both criteria, the exceptions being Inspec and JPTimes, where GPT-2 + BiLSTM-CRF performs the best out of all approaches according to F1@5. Another exception is the SemEval dataset, where the GPT-2 + BiLSTM-CRF is outperformed by TNT-KID by a large margin of about 12 percentage points. On the other hand, a GPT-2 model with a standard token classification head does not perform competitively on most datasets.

When it comes to the F1@5 measure, TNT-KID performs competitively on all the datasets. It outperforms all other algorithms on two datasets (KPTimes and DUC) and on average still performs the best out of all algorithms (see row \textit{average}). Nevertheless, the performance in terms of F1@5 is still noticeably worse than in terms of F1@10. The difference between TNT-KID and CatSeqD, which performs the best on four out of eight datasets in terms of F1@5, can be partially explained by the difference in training regimes and the fact that our system was designed to maximize recall (see Section \ref{sec:methodology}). Since our system generally detects more keywords than CatSeqD and CopyRNN, it tends to achieve better recall, which offers a better performance when up to 10 keywords need to be predicted. On the other hand, a more conservative system that generally predicts less keywords tends to achieve a better precision, which positively affects the F1 score in a setting where only up to 5 keywords need to be predicted. This phenomenon will be analysed in more detail in Section \ref{sec:error}, where we also discuss the very low results achieved by CopyRNN and CatSeqD on the DUC dataset.

When it comes to two other supervised approaches, KEA and Maui, they perform badly on all datasets they have been tested on and are outperformed by a large margin even by all unsupervised approaches. When we compare just unsupervised approaches, TextRank achieves by far the best results according to both measures on the Inspec dataset. This is the dataset with the on average shortest documents. On the other hand, TextRank performs uncompetitively in comparison to other unsupervised approaches on two datasets with much longer documents, KPTimes and JPTimes, where RaKUn and TfIdf are the best unsupervised approaches, respectively. Interestingly, it achieves the highest F@10 score out of all unsupervised keyword detectors on the DUC dataset, which also contains long documents. Perhaps this could be explained by the average number of present keywords, which is much higher for DUC-test (7.79) than for KPTimes-test (2.4) and JPTimes-test (3.86) datasets.  

Overall (see row \textit{average}), TNT-KID offers the most robust performance on the test datasets and is closely followed by GPT-2 + BiLSTM. CopyRNN and CatSeqD are very close to each other according to both criteria. Out of unsupervised approaches, on average all of them offer surprisingly similar performance. According to the F@10 score, YAKE on average works slightly better than the second ranked TextRank and also in general offers more steady performance, since it gives the most consistent results on a variety of different datasets. Similar could be said for RaKUn, the best ranked unsupervised algorithm according to the F@5 score.

Examples of the TNT-KID keyword detection are presented in the Appendix.

\section{Error analysis}
\label{sec:error}

In this Section we first analyse the reasons, why transformer based TNT-KID is capable of outperforming other state-of-the-art neural keyword detectors, which employ a generative model, by a large margin on some of the datasets. Secondly, we gather some insights into the inner workings of the TNT-KID by a visual analysis of the attention mechanism.
\subsection{Comparison between TNT-KID and CatSeqD}
\label{sec:compare_tnt_catseqd}

As was observed in Section \ref{sec-KEresults}, transformer based TNT-KID and GPT-2 + BiLSTM-CRF outperform generative models CatSeqD and CopyRNN by a large margin on the Inspec, JPTimes and DUC datasets. Here, we try to explain this discrepancy by focusing on the difference in performance between the best transformer based model, TNT-KID, and the best generative model, CatSeqD. The first hypothesis is connected with the statistical properties of the datasets used for training and testing, or more specifically, with the average number of keywords per document for each dataset. Note that CatSeqD is trained on the KP20k-train, when employed on the computer science domain, and on the KPTimes-train dataset, when employed on news. Table \ref{tbl:datasets} shows that both of these datasets do not contain many present keywords per document (KP20k-train 3.28 and KPTimes-train 2.38), therefore training the model on these datasets conditions it to be conservative in its predictions and to assign less keywords to each document than a more liberal TNT-KID. This gives the TNT-KID a competitive advantage on the datasets with more present keywords per document. 

\begin{figure}[b!]
    \centering
    \includegraphics[width = \linewidth]{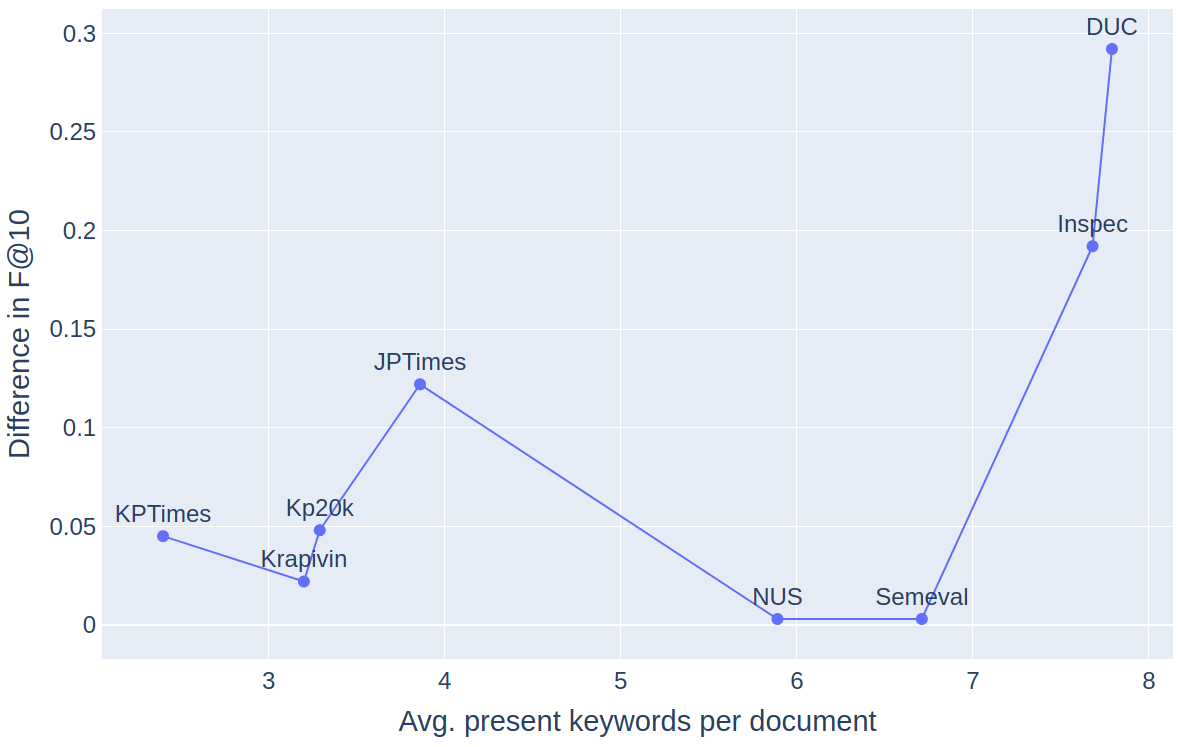}
    \caption{Relation between the average number of present keywords per document for each test dataset and the difference in performance ($F@10_{\textrm{TNT-KID}} - F@10_{\textrm{CatSeqD}}$).}
    \label{fig:present_kw}
\end{figure}

\begin{figure}[t!]
    \centering
    \includegraphics[width = \linewidth]{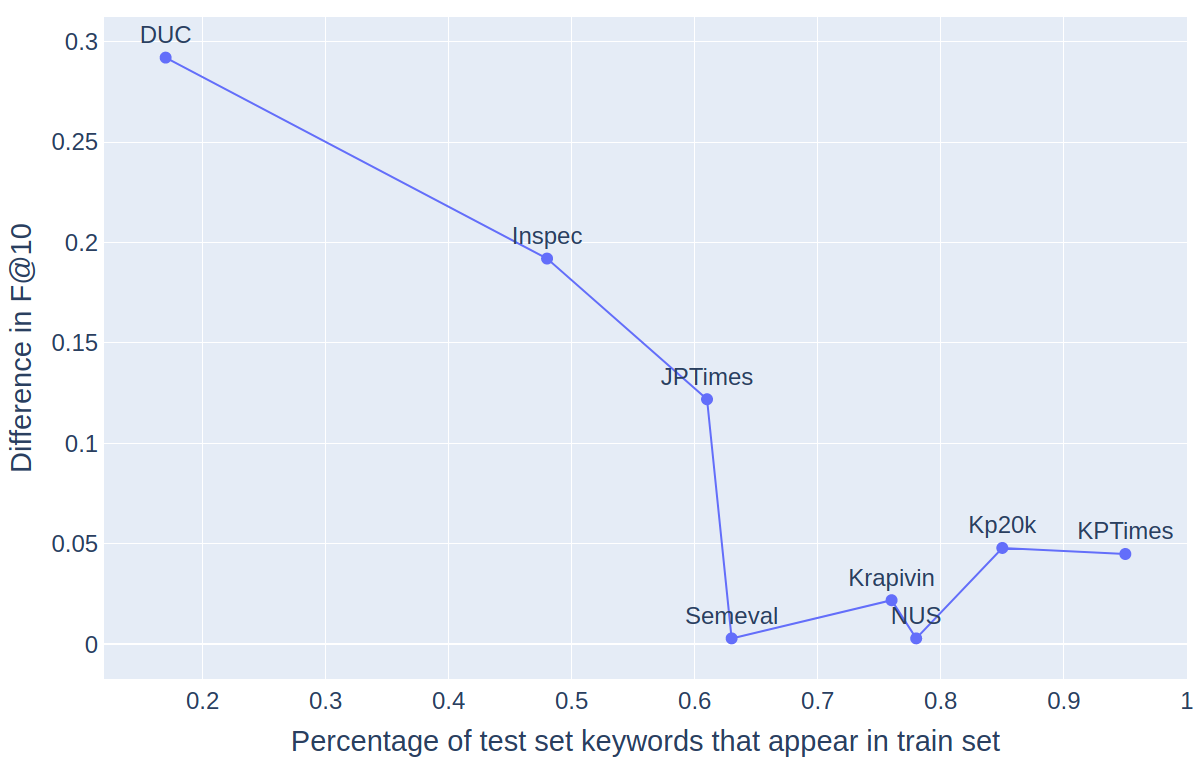}
    \caption{Relation between the percentage of keywords that appear in the train set for each test dataset and the difference in performance ($F@10_{\textrm{TNT-KID}} - F@10_{\textrm{CatSeqD}}$).}
    \label{fig:kw-overlap}
\end{figure}

Figure \ref{fig:present_kw} shows a correlation between the average number of present keywords per document for each dataset and the difference in performance in terms of F@10, measured as a difference between an F@10 score achieved by TNT-KID and an F@10 score achieved by CatSeqD. The difference in performance is the biggest for the DUC dataset (about 30 percentage points) that on average has the most keywords per document, 7.79, and second biggest for Inspec, in which an average document has 7.68 present keywords. 

The above hypothesis explains why CatSeqD offers competitive performance on the KP20k-test, Krapivin-test, NUS-test and KPTimes-test datasets with similar number of keywords per document than its two train sets but does not explain the competitive performance of CatSeqD on the SemEval test set that has 6.71 present keywords per document. Even more importantly, it does not explain the large difference in performance between TNT-KID and CatSeqD on the JPTimes-test. This suggests that there is another factor influencing the performance of some keyword detectors.

The second hypothesis suggests that the difference in performance could be explained by the difference in training regimes and the different tactics used for keyword detection by the two systems. While TNT-KID is fine-tuned on each of the datasets, no fine-tuning is conducted for CatSeqD that needs to rely only on the information obtained during training on the large KP20k-train and KPTimes-train datasets. This information seems sufficient when CatSeqD is tested on datasets that contain similar keywords than the train sets. On the other hand, this training regime does not work for datasets that have less overlapping keywords. 

Figure \ref{fig:kw-overlap} supports this hypothesis by showing strong correlation between the difference in performance in terms of F@10 and the percentage of keywords that appear both in the CatSeqD train sets (KP20k-train and KPTimes-train for computer science and news domain, respectively) and the test datasets. DUC and Inspec datasets have the smallest overlap, with only 17 percent of keywords in DUC appearing in the KPTimes-train and with 48 percent of keywords in Inspec appearing in the KP20k-train set. On the other hand, Krapivin, NUS, KP20k and KPTimes, the test sets on which CatSeqD performs more competitively, are the datasets with the biggest overlap, reaching up to 95 percent for KPTimes-test.

Figure \ref{fig:kw-overlap} also explains a relatively bad performance of CatSeqD on the JPTimes corpus (see Table \ref{tbl:results}) despite the smaller average number of keywords per document. Interestingly, despite the fact that no dataset specific fine-tuning for TNT-KID is conducted on the JPTimes corpus (since there is no validation set available, fine-tuning is conducted on the KPTimes-valid), TNT-KID manages to outperform CatSeqD on this dataset by about 12 percentage points. This suggests that a smaller keyword overlap between train and test sets has less of an influence on the TNT-KID and could be explained with the fact, that CatSeqD considers keyword extraction as a generation task and tries to generate a correct keyword sequence, while TNT-KID only needs to tag an already existing word sequence, which is an easier problem that perhaps requires less specific information gained during training.

According to the Figure \ref{fig:kw-overlap}, the SemEval test set is again somewhat of an outlier. Despite the keyword overlap that is quite similar to the one in the JPTimes test set and despite having a relatively large set of present keywords per document, CatSeqD still performs competitively on this corpus. This points to a hypothesis that there might be another unidentified factor, either negatively influencing the performance of TNT-KID and positively influencing the performance of CatSeqD, or the other way around.

\subsection{CatSeqD fine-tuning}
\label{sec:cat_seq_fine_tuning}

According to the results in Section \ref{sec-KEresults}, supervised approaches to the keyword extraction task tend to outperform unsupervised approaches, most likely due to their ability to adapt to the specifics of the syntax, semantics and keyword labeling regime of the specific corpus. On the other hand, the main disadvantage of most supervised approaches is that they require a large dataset with labeled keywords for training, which are scarce at least in some languages. In this paper we argue, that the main advantage of the proposed TNT-KID approach is, that due to its language model pretraining, the model only requires a small labeled dataset in order to fine-tune the language model for the keyword classification task. This fine-tuning  allows the model to adapt to each dataset and leads to a better performance of TNT-KID in comparison to CatSeqD, for which no fine-tuning was conducted.

\begin{figure}[t!]
    \centering
    \includegraphics[width = \linewidth]{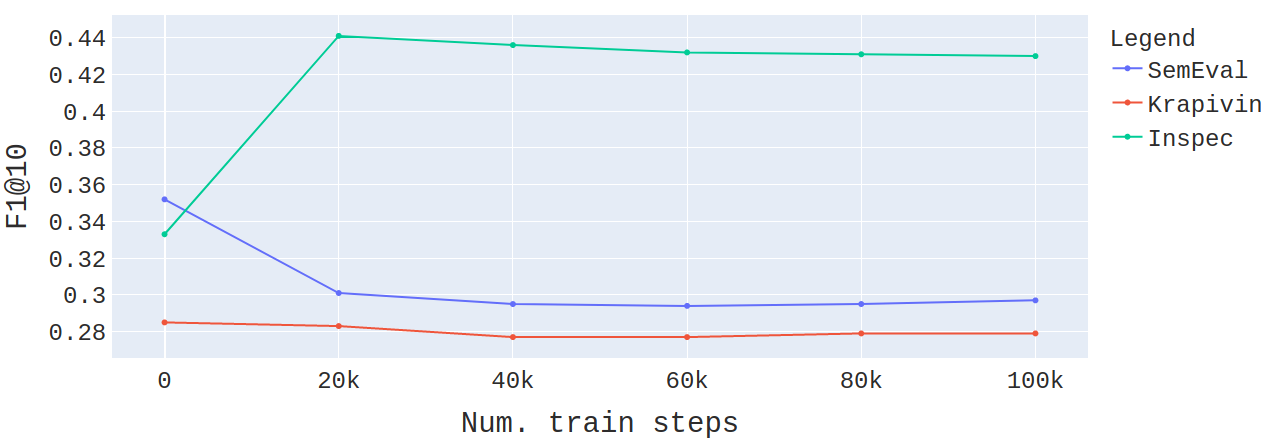}
    \caption{Performance of the KP20k trained CatSeqD model fine-tuned on SemEval, Krapivin and Inspec validation sets and tested on the corresponding test sets, in correlation with the length of the fine tuning in terms of number of train steps. Zero train steps means that the model was not fine-tuned.}
    \label{fig:catseqd-fine-tuning}
\end{figure}

Even though no fine-tuning was conducted in the original CatSeqD study \citep{yuan2019one}, one might hypothesise that the performance of CatSeqD could be further improved if the model would be fine-tuned on each dataset, same as TNT-KID. To test this hypothesis, we take the CatSeqD model trained on KP20k, conduct additional training on the SemEval, Krapivin and Inspec validation sets (i.e., all datasets besides KP20k and KPTimes with a validation set), and test these fine-tuned models on the corresponding test sets. Fine-tuning was conducted for up to 100.000 train steps\footnote{Same hyperparameters that were used for KP20k training in the original paper \citep{yuan2019one} were used for fine-tuning.} and the results are presented in Figure \ref{fig:catseqd-fine-tuning}.

Only on one of the three datasets, the Inspec test set, the performance can be improved by additional fine-tuning. Though the improvement on the Inspec test set of about 10 percentage points (from 33.5\% to 44\%) in terms of F1@10 is quite substantial, the model still performs worse than TNT-KID, which achieves F1@10 of 52.5\%. The improvement is most likely connected with the fact that the Inspec test set contains more keywords that do not appear in the KP20k than SemEval and Krapivin test sets (see Figure \ref{fig:kw-overlap}). Inspec test set also contains more keywords per document than the other two test sets (7.68 present keyword on average, in comparison to 6.71 present keywords per document in the SemEval test set and 3.2 in the Krapivin test set). Since the KP20k train set on average contains only 3.29 present keywords per document, the fine-tuning on the Inspec dataset most likely also adapts the classifier to a more liberal keyword labeling regime.

On the other hand, fine-tuning does not improve the performance on the Krapivin and SemEval datasets. While there is no difference between the fine-tuned and original model on the Krapivin test set, fine-tuning negatively affects the performance of the model on the SemEval dataset. The F1@10 score drops from about 35\% to about 30\% after 20.000 train steps. Further fine-tuning does not have any effect on the performance. The hypothesis is, that this drop in performance is somewhat correlated with the size of the SemEval validation set, which is much smaller (it contains only 144 documents) than  Inspec and Krapivin validation sets (containing 1500 and 1844 documents, respectively), and this causes the model to overfit. Further tests would however need to be conducted to confirm or deny this hypothesis.

Overall, 20.000 train steps seem to be enough for model adaptation in each case, since the results show that additional fine-tuning does not have any influence on the performance.

\subsection{Dissecting the attention space}

One of the advantages of the transformer architecture is its employment of the attention mechanism, that can be analysed and visualized, offering valuable insights into inner workings of the system and enabling interpretation of how the neural net tackles the keyword identification task. The TNT-KID attention mechanism consists of multiple attention heads \citep{vaswani2017attention} -- square matrices linking pairs of tokens within a given text -- and we explored how this (activated) weight space can be further inspected via visualization and used for interpretation.

\begin{figure}[b!]
    \centering
    \includegraphics[width = .85\linewidth]{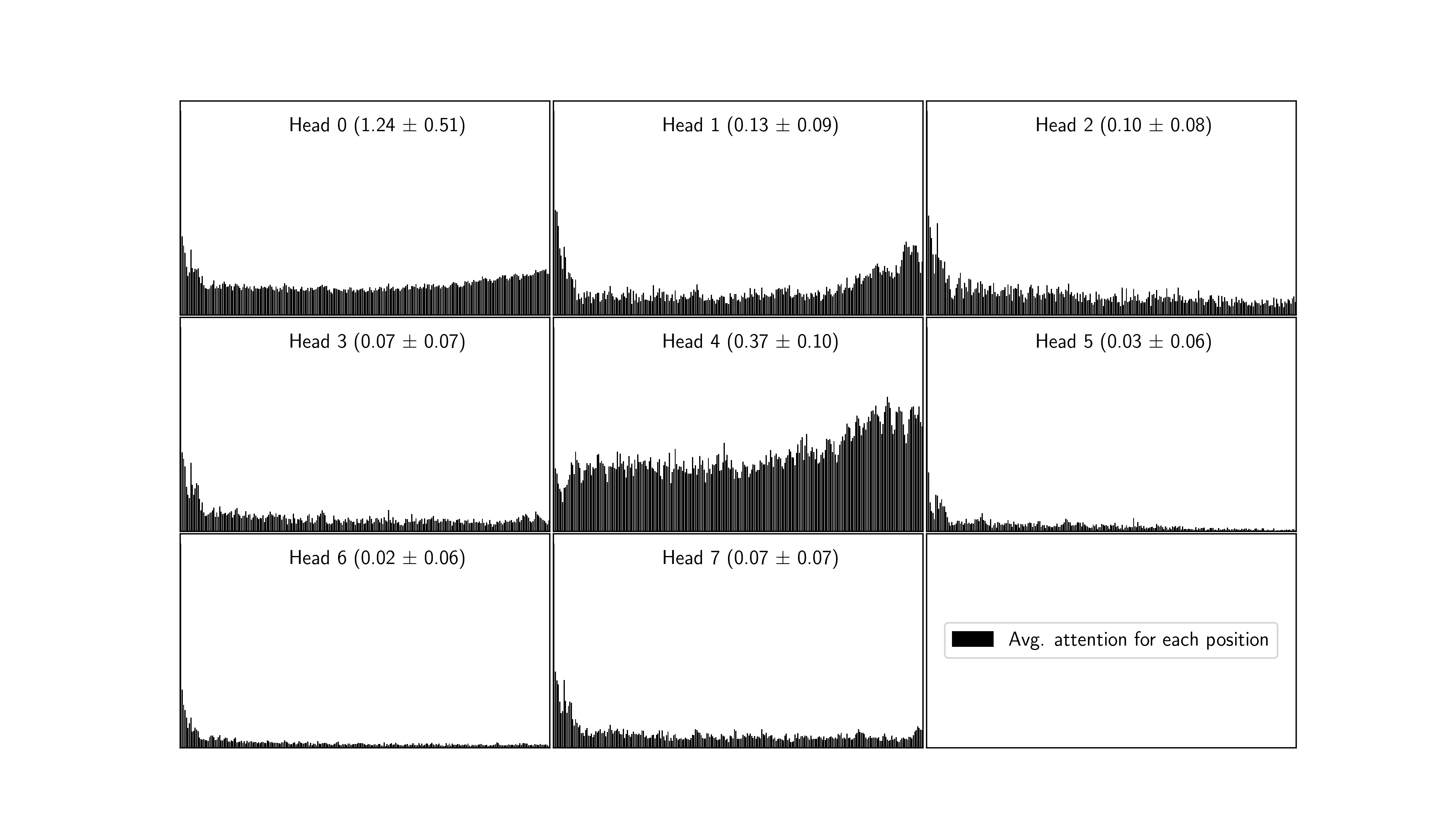}
    \caption{Average attention for each token position in the SemEval corpus across eight attention heads. Distinct peaks can be observed for tokens appearing at the beginning of the document in all but one out of eight attention heads.}
    \label{fig:heads-self}

    \centering
    \includegraphics[width = .85\linewidth]{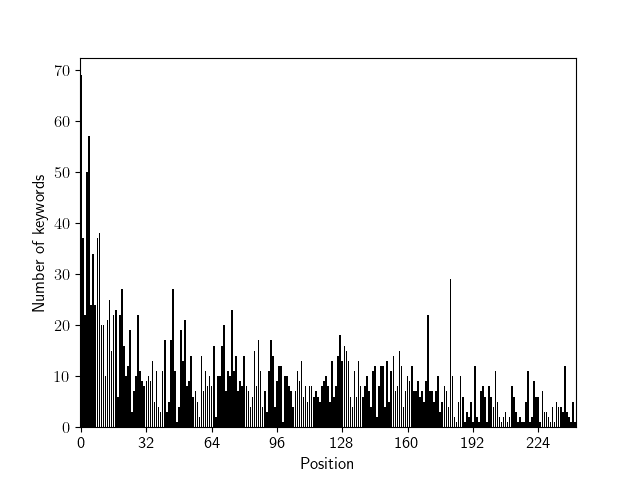}
    \caption{Number of keywords for each token position in the SemEval corpus. Distinct peaks can be observed for positions at the beginning of the document.}
    \label{fig:semeval-kw-position}
\end{figure}

While square attention matrices show importance of the correlations between all tokens in the document for a keyword identification task, we focused only on the diagonals of the matrices, which indicate how much attention the model pays to the  ``correlation'' a specific word has with itself, i.e., how important is a specific word for the classification of a specific token as either being a keyword or not. We extracted these diagonal attention scores for eight attention heads of the last out of eight encoders, for each of the documents in the SemEval-test and averaged the scores across an entire dataset by summing together scores belonging to the same position in each head and dividing this sum with the number of documents. Figure~\ref{fig:heads-self} shows the average attention score of each of the eight attention heads for each token position. While there are distinct differences between heads, a distinct peak at the beginning of the attention graph can be observed for all heads but one (head 4), which means that heads generally pay more attention to the tokens at the beginning of the document. This suggests that the system has learned that tokens appearing at the beginning of the document are more likely to be keywords (Figure \ref{fig:semeval-kw-position} shows the actual keyword count for each position in the SemEval corpus) and once again shows the importance of positional information for the task of keyword identification.

\begin{figure}[b!]
    \centering
    \includegraphics[width = \linewidth]{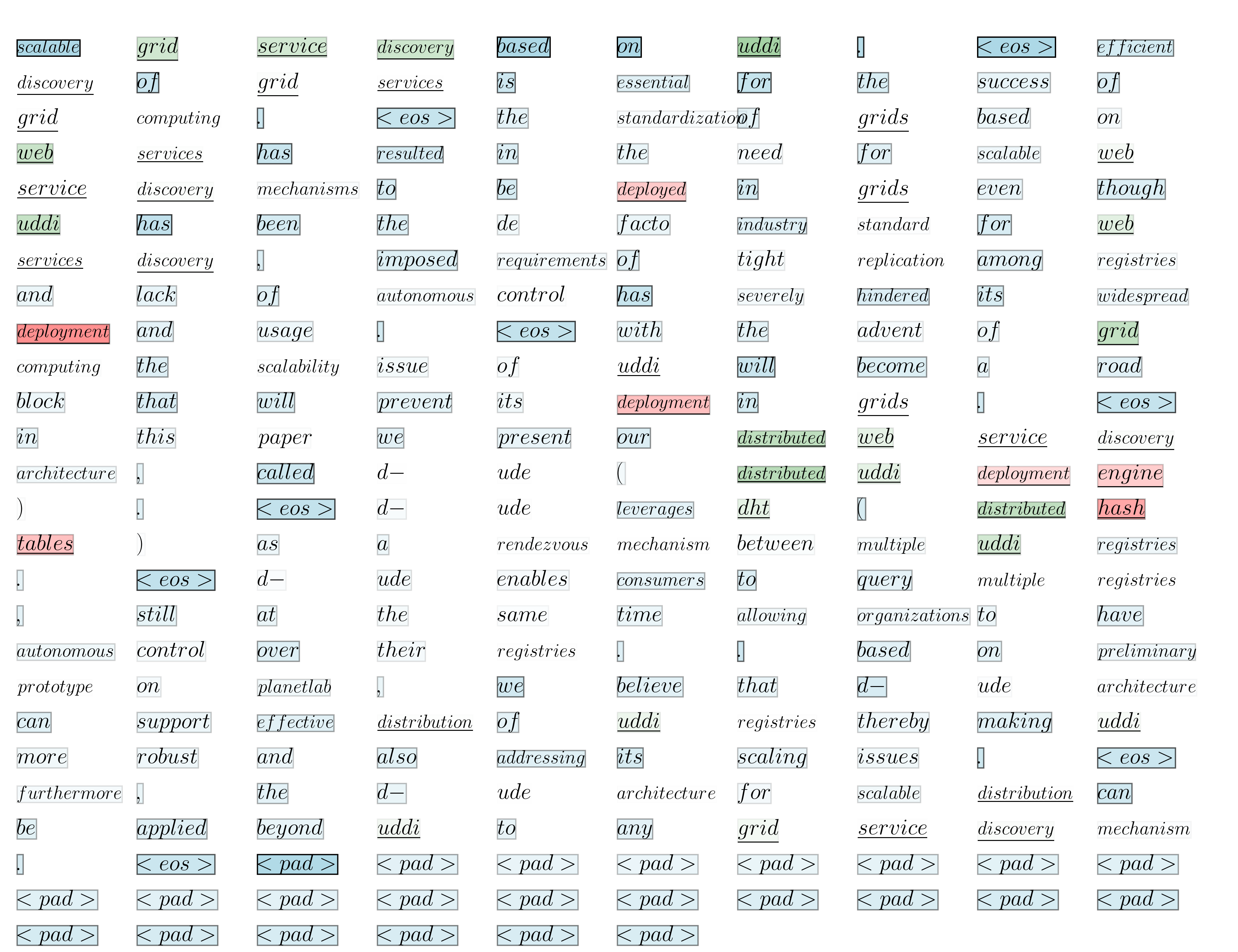}
    \caption{Attention-colored tokens. Green ones were correctly identified as keywords, red ones were incorrectly identified as keywords and less transparency indicates stronger attention for the token. Underlined words in italic were identified as keywords by the system.}
    \label{fig:attention-viz}
\end{figure}

Another insight into how the system works can be gained by analysing how much attention was paid to each individual token in each document. Figure~\ref{fig:attention-viz} displays attentions for individual tokens, as well as marks them based on predictions for an example document from the SemEval-test. Green tokens were correctly identified as keywords, red tokens were incorrectly identified as keywords and less transparency (more colour) indicates that a specific token received more attention from the classifier.

Figure \ref{fig:attention-viz} shows that, at least for this specific document, tokens that were either correctly or incorrectly classified as keywords did receive more attention than an average token. There are also some tokens that received a lot of attention and were not classified as keywords, e.g., \textit{eos} (end of sentence signs) and \textit{pad} (padding) signs, and also words like \textit{of, is, we, etc.}. Another interesting thing to notice is the fact, that the amount of attention associated with individual tokens that appear more than once in the document varies and is somewhat dependent on the position of the token. \footnote{Note that Figure \ref{fig:attention-viz} is just a motivating example. A more thorough statistical analysis of much more than just one document would be required in order to draw proper conclusions about the behavior of the attention mechanism during keyword identification.} 


\section{Ablation study}
\label{sec:ablation}

In this section we explore the influence of several technique choices and building blocks of the keyword extraction workflow on the overall performance of the model:

\begin{itemize}
    \item \textbf{Language model pretraining}; assessment whether pretraining positively affects the performance of the keyword extraction and if the improvements are dataset or domain specific.
    \item \textbf{Choice of pretraining regime}; comparison of two pretraining objectives, autoregressive language modelling and masked language modelling described in Section \ref{sec:transfer}.
    \item \textbf{Choice of input tokenization scheme}; comparison of two tokenization schemes, word tokenization and Sentencepiece \citep{kudo2018sentencepiece} byte-pair encoding. 
    \item \textbf{Part-of-speech(POS) tags}; assessment whether adding POS tags as an additional input improves the performance of the model.
    \item \textbf{Transformer architecture adaptations}; as was explained in Section \ref{sec:architecture}, in the fine-tuning stage we add an additional BiLSTM encoder to the output of the transformer encoder. We also experiment with the addition of the BiLSTM+CRF token classification head on top of the model, as was proposed in \cite{sahrawat2019keyphrase} and described in Section \ref{sec:architecture}. Here we assess the influence of these additions on the performance of the model. 
\end{itemize}

Table \ref{tbl:ablation} presents results on all datasets for several versions of the model, a model with no language model pretraining (\textit{nolm}), a model pretrained with an autoregressive language model objective (\textit{lm}), a model pretrained with a masked language model objective (\textit{maskedlm}), a model pretrained with an autoregressive language model objective and leveraging byte-pair encoding tokenization scheme (\textit{lm+bpe}), a model pretrained with an autoregressive language model objective and leveraging additional POS tag sequence input (\textit{lm+pos}), a model pretrained with an autoregressive language model objective and a BiLSTM encoder (\textit{lm+rnn}), a model pretrained with an autoregressive language model objective leveraging byte-pair encoding tokenization scheme and a BiLSTM encoder (\textit{lm+bpe+rnn}), and a model pretrained with an autoregressive language model objective leveraging byte-pair encoding tokenization scheme and a BiLSTM+CRF token classification head (\textit{lm+bpe+crf}) .

\begin{table}[t!]
    \centering
        \caption{Results of the ablation study. Column \textit{lm+bpe+rnn} represents the results for the model that was used for comparison with other methods from the related work in Section \ref{sec-KEresults}.}
    \resizebox{0.99\textwidth}{!}{
    \begin{tabular}{l|cccccccc}
& nolm & lm & maskedlm & lm+bpe & lm+pos & lm+rnn & \textbf{lm+bpe+rnn} & lm+bpe+crf\\\hline
& \multicolumn{8}{c}{\textbf{KP20k}}\\
F1@5 & 0.2544 & 0.2922 & 0.2476 & 0.2958 & 0.3003 & 0.3349 & 0.3418 & \textbf{0.3478}\\
F1@10 & 0.2304 & 0.2836 & 0.2313 & 0.2941 & 0.2986 & 0.3382 & 0.3457 & \textbf{0.3521}\\\hline
& \multicolumn{8}{c}{\textbf{Inspec}}\\
F1@5 & 0.2868 & 0.4099 & 0.2875 & 0.4255 & 0.4136 & \textbf{0.4506} & 0.4471 & 0.4463\\
F1@10 & 0.3636 & 0.4994 & 0.3704 & 0.4871 & 0.5012 & \textbf{0.5253} & 0.5252 & 0.5147\\\hline
& \multicolumn{8}{c}{\textbf{Krapivin}}\\
F1@5 & 0.1919 & 0.2277 & 0.2046 & 0.2879 & 0.2494 & 0.3088 & 0.3009 & \textbf{0.3142}\\
F1@10 & 0.1904 & 0.2314 & 0.2029 & 0.2884 & 0.2555 & 0.3164 & 0.3070 & \textbf{0.3178}\\\hline
& \multicolumn{8}{c}{\textbf{NUS}}\\
F1@5 & 0.1909 & 0.3319 & 0.2372 & 0.3352 & 0.3339 & 0.3419 & \textbf{0.3502} & 0.3371\\
F1@10 & 0.1902 & 0.3492 & 0.2552 & 0.3586 & 0.3518 & 0.3626 & \textbf{0.3686} & 0.3658\\\hline
& \multicolumn{8}{c}{\textbf{SemEval}}\\
F1@5 & 0.1671 & 0.3070 & 0.1842 & 0.2462 & 0.2780 & 0.2696 & \textbf{0.2921} & 0.2524\\
F1@10 & 0.1950 & 0.3469 & 0.2528 & 0.2913 & 0.3426 & 0.3303 & \textbf{0.3552} & 0.3007\\\hline
& \multicolumn{8}{c}{\textbf{KPTimes}}\\
F1@5 & 0.2864 & 0.4242 & 0.3052 & 0.4211 & 0.4306 & 0.4627 & \textbf{0.4691} & 0.4408\\
F1@10 & 0.2760 & 0.4208 & 0.3017 & 0.4208 & 0.4300 & 0.4609 & \textbf{0.4693} & 0.4413\\\hline
& \multicolumn{8}{c}{\textbf{JPTimes}}\\
F1@5 & 0.2490 & 0.3305 & 0.2644 & 0.3341 & 0.3359 & \textbf{0.3790} & 0.3570 & 0.3357\\
F1@10 & 0.2478 & 0.3344 & 0.2705 & 0.3373 & 0.3402 & \textbf{0.3823} & 0.3596 & 0.3372\\\hline
& \multicolumn{8}{c}{\textbf{DUC}}\\
F1@5 & 0.1951 & 0.2848 & 0.1523 & 0.2759 & 0.2918 & 0.3003 & \textbf{0.3115} & 0.2943\\
F1@10 & 0.2265 & 0.3340 & 0.1979 & 0.3213 & 0.3386 & 0.3432 & \textbf{0.3551} & 0.3342\\\hline\hline
& \multicolumn{8}{c}{\textbf{Average}}\\
F@5 & 0.2277 & 0.3260 & 0.2354 & 0.3277 & 0.3292 & 0.3560 & \textbf{0.3587} & 0.3461\\
F@10 & 0.2400 & 0.3500 & 0.2603 & 0.3499 & 0.3573 & 0.3824 & \textbf{0.3857} & 0.3705\\\hline
\end{tabular}
    }
\label{tbl:ablation}
\end{table}

On average (see last two rows in Table \ref{tbl:ablation}), by far the biggest boost in performance is gained by employing the autoregressive language model pretraining (column \textit{lm}), improving the F@5 score by about 10 percentage points and the F@10 score by 11 percentage points in comparison to no language model pretraining (column \textit{nolm}). As expected, the improvements are substantial on three smallest corpora, which by themselves do not contain enough text for the model to obtain sufficient syntactic and semantic knowledge. The largest gains are achieved on the NUS test set, where almost an 84 percent improvement in terms of the F@10 score can be observed, and on the SemEval test set, where the improvement of 78 percent in terms of F@5 can be observed. We also observe about a 47 percent improvement in terms of F@10 on the DUC test set. Not surprisingly, for the KP20k dataset, which has a relatively large validation set used for fine-tuning, we can observe a much smaller improvement of about 23 percent in terms of F@10. On the other hand, we observe a substantial improvement of roughly 50 percent in terms of both F@5 and F@10 on the KPTimes test set, even though the KPTimes validation set used for fine-tuning is the same size as KP20k validation set. This means that in the language modelling phase the model still manages to obtain knowledge that is not reachable in the fine-tuning phase and can perhaps be partially explained by the fact that all documents are truncated into 256 tokens long sequences in the fine-tuning phase. The KPTimes-valid dataset, used both for language modelling and fine-tuning, has on average 784.65 tokens per document, which means that more than half of the document's text is discarded during the fine-tuning phase. This is not the case in the language modelling phase, where all of the text is leveraged.

On the other hand, using the masked language modelling pretraining (column \textit{maskedlm}) objective on average yields only a somewhat negligible improvement of about 0.8 percentage point in terms of F@5 score and a slightly bigger improvement of about 2 percentage points in terms of F@10 score in comparison to no language model pretraining. It does however improve the performance on the two smallest datasets, NUS (by about 6.5 percentage points in terms of F1@10) and SemEval (by about 6 percentage points in terms of F1@10). The large discrepancy in performance between the two different language model objectives can be partially explained by the sizes of the pretraining corpora. By using autoregressive language modelling, the model learns to predict the next word probability distribution for each sequence in the corpus. By using the masked language modelling objective, 15 percent of the words in the corpus are randomly masked and used as targets for which the word probability distributions need to be predicted from the surrounding context. Even though each training epoch a different set of words is randomly masked, it is quite possible, that some words are never masked due to small sizes of the corpora and since we only train the model for up to 10 epochs. 

Results show that adding POS tags as an additional input (column \textit{lm+pos}) leads to only marginal performance improvements. Some previous studies suggest that transformer based models that employ transfer learning already capture sufficient amount of syntactic and other information about the composition of the text \citep{jawahar2019does}. Our results therefore support the hypothesis that additional POS tag inputs are somewhat unnecessary in the transfer learning setting but additional experiments would be needed to determine whether this is task/language specific or not. 

Another adaptation that does not lead to any significant improvements when compared to the column \textit{lm} is the usage of the byte-pair encoding scheme (column \textit{lm+bpe}). The initial hypothesis that motivated the usage of byte-pair encoding was that it might help the model's performance by introducing some knowledge about the word composition and by enabling the model to better understand that different forms of the word can represent the same meaning. However, the usage of byte-pair encoding might on the other hand also negatively affect the performance, since splitting up words inside a specific keyphrase would make these keyphrases longer in terms of number of words and detecting a longer continuous word sequence as a keyword might represent a harder problem for the model than detecting a shorter one. Nevertheless, usage of byte-pair encoding does have an additional positive effect of drastically reducing the vocabulary of the model (e.g., for computer science articles, this means a reduction from about 250.000 tokens to about 30.000) and with it also the number of parameters in the model (from about 290 million to about 70 million). 

Adding an additional BiLSTM encoder in the fine-tuning stage of a pretrained model (column \textit{lm+rnn}) leads to consistent improvements on almost all datasets and to an average improvement of about 3 percentage points in terms of both F@5 and F@10 scores. This confirms the findings from the related work that recurrent neural networks work well for the keyword detection task and also explains why a majority of state-of-the-art keyword detection systems leverage recurrent layers. 

We also present results for a model in which we employed autoregressive language model pretraining, used byte-pair encoding scheme and added a BiLSTM encoder (column \textit{lm+bpe+rnn}) that was used for comparison with other methods from the related work in Section \ref{sec-KEresults}, and results for the approach proposed by \cite{sahrawat2019keyphrase}, where a BiLSTM+CRF token classification head is added on top of the transformer encoder, that employs byte-pair encoding scheme and autoregressive language model pretraining (column \textit{lm+bpe+crf}). The BiLSTM+CRF performs quite well, outperforming all other configurations on two (KP20k and Krapivin) datasets. On average it however still performs by more than 1 percentage point worse than both configurations employing an added BiLSTM encoder.

\section{Conclusion and future work}
\label{sec:conclusion}

In this research we have presented TNT-KID, a novel transformer based neural tagger for keyword identification that leverages a transfer learning approach to enable robust keyword identification on a number of datasets. The presented results show that the proposed model offers a robust performance across a variety of datasets with manually labeled keywords from two different domains. By exploring the differences in performance between our model and the best performing generative model from the related work, CatSeqD by \cite{yuan2019one}, we manage to pinpoint strengths and weaknesses of each model and therefore enable a potential user to choose the approach most suitable for the task at hand. By visualizing the attention mechanism of the model, we try to interpret classification decisions of the neural network and show that efficient modelling of positional information is essential in the keyword detection task. Finally, we propose an ablation study which shows how specific components of the keyword extraction workflow influence the overall performance of the model. 

The biggest advantage of supervised approaches to keyword extraction task is their ability to adapt to the specifics of the syntax, semantics, content, genre and keyword tagging regime of the specific corpus. Our results show that this offers a significant performance boost and state-of-the-art supervised approaches outperform state-of-the-art unsupervised approaches on the majority of datasets. On the other hand, the ability of the supervised models to adapt might become limited in cases when the train dataset is not sufficiently similar to the dataset on which keyword detection needs to be performed. This can clearly be seen on the DUC dataset, in which only about 17 percent of keywords also appear in the KPTimes train set, used for training the generative CopyRNN and CatSeqD models. Here, these two state-of-the-art models perform the worst of all the models tested and as is shown in Section \ref{sec:cat_seq_fine_tuning}, this \textit{keywordinees} generalization problem can not be overcome by simply fine-tuning these state-of-the-art systems on each specific dataset.

On the other hand, TNT-KID bypasses the generalization problem by allowing fine-tuning on very small datasets. Nevertheless, the results on the JPTimes corpus suggest that it also generalizes better than CopyRNN and CatSeqD. Even though all three algorithms are trained on the KPTimes dataset (since JPTimes corpus does not have a validation set)\footnote{Note that TNT-KID is trained on the validation set, while CopyRNN and CatSeqD are trained on the much larger train set.}, TNT-KID manages to outperform the other two by about 10 percentage points according to the F1@10 and F1@5 criteria despite the discrepancy between train and test set keywords. As already mentioned in Section \ref{sec:compare_tnt_catseqd}, this can be partially explained by the difference in approaches used by the models and the fact that keyword generation is a much harder task than keyword tagging. For keyword generation task to be successful, seeing a sequence that needs to be generated in advance, during training, is perhaps more important, than for a much simpler task of keyword tagging, where a model only needs to decide if a word is a keyword or not. Even though the keyword generators try to ease the task by employing a copying mechanism \citep{gu2016incorporating}, the experiments suggest that generalizing \textit{keywordinees} to unseen word sequences still represent a bigger challenge for these models than for TNT-KID. 

While the conducted experiments suggest that TNT-KID works better than other neural networks in a setting where previously unseen keywords (i.e., keywords not present in the training set) need to be detected, further experiments need to be devised to evaluate the competitiveness of TNT-KID in a cross-domain setting when compared to unsupervised approaches. Therefore, in order to determine if the model's internal representation of \textit{keywordiness} is general enough to be transferable across different domains, in the future we also plan to conduct some cross-domain experiments. 

Another aspect worth mentioning is the evaluation regime and how it affects the comparison between the models. By fine-tuning the model on each dataset, the TNT-KID model learns the optimal number of keywords to predict for each specific dataset. This number is in general slightly above the average number of present keywords in the dataset, since the loss function was adapted to maximize recall (see Section \ref{sec:methodology}). On the other hand, CatSeqD and CopyRNN are only trained on the KP20k-train and KPTimes-train datasets that have less present keywords than a majority of test datasets. This means our system on average predicts more keywords per document than these two systems, which negatively affects the precision of the proposed system in comparison to CatSeqD and CopyRNN, especially at smaller k values. On the other hand, predicting less keywords hurts recall, especially on datasets where documents have on average more keywords. As already mentioned in Section \ref{sec:ablation}, this explains why our model compares better to other systems in terms of F@10 than in terms of F@5 and also raises a question how biased these measures of performance actually are. Therefore, in the future we plan to use other performance measures to compare our model to others.

Overall, the differences in training and prediction regimes between TNT-KID and other neural models imply that the choice of a network is somewhat dependent on the use-case. If a large training dataset of an appropriate genre with manually labeled keywords is available and if the system does not need to predict many keywords, then CatSeqD is most likely the best choice, even though TNT-KID shows competitive performance on a large majority of datasets. On the other hand, if only a relatively small train set is available and it is preferable to predict a larger number of keywords, then the results of this study suggest that TNT-KID is most likely a better choice.

The conducted study also indicates that the adaptation of the transformer architecture and the training regime for the task at hand can lead to improvements in keyword detection. Both TNT-KID and a pretrained GPT-2 model with a BiLSTM + CRF token classification head manage to outperform the unmodified GPT-2 with a default token classification head by a large margin. Even more, TNT-KID manages to outperform both, the pretrained GPT-2 and the GPT-2 with BiLSTM + CRF, even though it employs only 8 attention layers, 8 attention heads and an embedding size of 512 instead of the standard 12 attention layers, 12 attention heads and an embeddings size of 768, which the pretrained GPT-2 employs. The model on the other hand does employ an additional BiLSTM encoder during the classification phase, which makes it slower than the unmodified GPT-2 but still faster than the GPT-2 with the BiLSTM + CRF token classification head that employs a computationally demanding CRF layer.

The ablation study clearly shows that the employment of transfer learning is by far the biggest contributor to the overall performance of the system. Surprisingly, there is a very noticeable difference between performances of two distinct pretraining regimes, autoregressive language modelling and masked language modelling in the proposed setting with limited textual resources. Perhaps a masked language modelling objective regime could be somewhat improved by a more sophisticated masking strategy that would not just randomly mask 15 percent of the words but would employ a more fine-grained entity-level masking and phrase-level masking, similar as in \cite{sun2019ernie}. This and other pretraining learning objectives will be explored in future work.

In the future we also plan to expand the set of experiments in order to also cover other languages and domains. Since TNT-KID does not require a lot of manually labeled data for fine-tuning and only a relatively small domain specific corpus for pretraining, the system is already fairly transferable to other languages and domains, even to low resource ones. Deploying the system to a morphologically richer language than English and conducting an ablation study in that setting would also allow us to see, whether byte-pair encoding and the additional POS tag sequence input would lead to bigger performance boosts on languages other than English. 

Finally, another line of research we plan to investigate is a cross-lingual keyword detection. The idea is to pretrain the model on a multilingual corpus, fine-tune it on one language and then conduct zero-shot cross-lingual testing of the model on the second language. Achieving a satisfactory performance in this setting would make the model transferable even to languages with no manually labeled resources.    

\begin{acknowledgements}
This paper is supported by European Union’s Horizon 2020 research and  innovation programme under grant agreement No. 825153, project EMBEDDIA (Cross-Lingual Embeddings for Less-Represented Languages in European News Media). The authors acknowledge also the financial support from the Slovenian Research Agency for research core funding for the programme Knowledge Technologies (No. P2-0103) and the project TermFrame - Terminology and Knowledge Frames across Languages (No. J6-9372).
\end{acknowledgements}

%
%

\bibliographystyle{authordate1}

\bibliography{bibliography}

\section*{Appendix: examples of keyword identification}

\doc{Quantum market games. We propose a quantum-like description of markets and economics. The approach has roots in the recently developed quantum game theory"
}

\pred{markets, quantum market games, quantum game theory, economics, quantum like description}
\true{economics, quantum market games, quantum game theory}

\doc{Revenue Analysis of a Family of Ranking Rules for Keyword Auctions. Keyword auctions lie at the core of the business models of today's leading search engines. Advertisers bid for placement alongside search results, and are charged for clicks on their ads. Advertisers are typically ranked according to a score that takes into account their bids and potential clickthrough rates. We consider a family of ranking rules that contains those typically used to model Yahoo! and Google's auction designs as special cases. We find that in general neither of these is necessarily revenue-optimal in equilibrium, and that the choice of ranking rule can be guided by considering the correlation between bidders' values and click-through rates. We propose a simple approach to determine a revenue-optimal ranking rule within our family, taking into account effects on advertiser satisfaction and user experience. We illustrate the approach using Monte-Carlo simulations based on distributions fitted to Yahoo! bid and click-through rate data for a high-volume keyword.}

\pred{ranked, auction, clicks, keyword auctions, keyword, revenue, clickthrough, ranking rules, click through rates, bids}
\true{revenue optimal ranking, ranking rule, revenue, advertisement, keyword auction, search engine}

\doc{Profile-driven instruction level parallel scheduling with application to super blocks. Code scheduling to exploit instruction level parallelism (ILP) is a critical problem in compiler optimization research in light of the increased use of long-instruction-word machines. Unfortunately optimum scheduling is computationally intractable, and one must resort to carefully crafted heuristics in practice. If the scope of application of a scheduling heuristic is limited to basic blocks, considerable performance loss may be incurred at block boundaries. To overcome this obstacle, basic blocks can be coalesced across branches to form larger regions such as super blocks. In the literature, these regions are typically scheduled using algorithms that are either oblivious to profile information (under the assumption that the process of forming the region has fully utilized the profile information), or use the profile information as an addendum to classical scheduling techniques. We believe that even for the simple case of linear code regions such as super blocks, additional performance improvement can be gained by utilizing the profile information in scheduling as well. We propose a general paradigm for converting any profile-insensitive list scheduler to a profile-sensitive scheduler. Our technique is developed via a theoretical analysis of a simplified abstract model of the general problem of profile-driven scheduling over any acyclic code region, yielding a scoring measure for ranking branch instructions.
}

\pred{scheduling;instruction level parallel scheduling;instruction level parallelism;profile;list scheduler;code scheduling;long instruction word machines;profile driven scheduling}
\true{long instruction word machines, scheduling heuristic, compiler optimization, optimum scheduling, abstract model, ranking branch instructions, profile driven instruction level parallel scheduling, profile sensitive scheduler, linear code regions, code scheduling}

\doc{40 Years After War, Israel Weighs Remaining Risks. JERUSALEM It was 1 p.m. on Saturday, Oct. 6, 1973, the day of Yom Kippur, 
the holiest in the Jewish calendar, and Israel's military intelligence chief, Maj. Gen. Eli Zeira, had called in the country's top military journalists for an urgent briefing. 
He told us that war would break out at sundown, about 6 p.m., said Nachman Shai, who was then the military affairs correspondent for Israel's public television channel and is now a Labor member of Parliament. 
Forty minutes later he was handed a note and said, Gentlemen, the war broke out, and he left the room. Moments before that note arrived, according to someone else who was at that meeting, 
General Zeira had been carefully peeling almonds in a bowl of ice water. The coordinated attack by Egypt and Syria, which were bent on regaining strategic territories and pride lost to Israel in the 1967 war, 
surprised and traumatized Israel. For months, its leaders misread the signals and wrongly assumed that Israel's enemies were not ready to attack. Even in those final hours, when the signs were unmistakable that a conflict was imminent, 
Israel was misled by false intelligence about when it would start. As the country's military hurriedly called up its reserves and struggled for days to contain, then repel, the joint assault, a sense of doom spread through the country. 
Many feared a catastrophe. Forty years later, Israel is again marking Yom Kippur, which falls on Saturday, the anniversary of the 1973 war according to the Hebrew calendar. This year the holy day comes in the shadow of new regional 
tensions and a decision by the United States to opt, at least for now, for a diplomatic agreement rather than a military strike against Syria in response to a deadly chemical weapons attack in the Damascus suburbs on Aug. 21. 
Israeli newspapers and television and radio programs have been filled with recollections of the 1973 war, even as the country's leaders have insisted that the probability of any new Israeli entanglement remains 
low and that the population should carry on as normal. For some people here, though, the echoes of the past have stirred latent questions about the reliability of intelligence assessments and the risks of another surprise attack. 
Any Israeli with a 40-year perspective will have doubts,said Mr. Shai, who was the military's chief spokesman during the Persian Gulf War of 1991, when Israelis huddled in sealed rooms and donned gas masks, 
shocked once again as Iraqi Scud missiles slammed into the heart of Tel Aviv. Coming after the euphoria of Israel's victory in the 1967 war, when six days of fighting against the Egyptian, Jordanian and Syrian Armies left Israel 
in control of the Sinai Peninsula, the West Bank, Gaza, East Jerusalem and the Golan Heights, the conflicts of 1973, 1991 and later years have scarred the national psyche. But several former security officials and analysts said that 
while the risks now may be similar to those of past years in some respects, there are also major differences. In 1991, for example, the United States responded to the Iraqi attack by hastily redeploying some Patriot 
antimissile batteries to Israel from Europe, but the batteries failed to intercept a single Iraqi Scud, tracking them instead and following them to the ground with a thud. Since then, Israel and the United States have invested billions of dollars 
in Israel's air defenses, with the Arrow, Patriot and Iron Dome systems now honed to intercept short-, medium- and longer-range rockets and missiles. Israelis, conditioned by subsequent conflicts with Hezbollah in Lebanon and
Hamas in Gaza and by numerous domestic drills, have become accustomed to the wail of sirens and the idea of rocket attacks. But the country is less prepared for a major chemical attack, even though chemical
weapons were used across its northern frontier, in Syria, less than a month ago, which led to a run on gas masks at distribution centers here. In what some people see as a new sign of government complacency at
best and downright failure at worst, officials say there are enough protective kits for only 60 percent of the population, and supplies are dwindling fast. Israeli security assessments rate the probability of any attack on Israel as low,
and the chances of a chemical attack as next to zero. In 1973, the failure of intelligence assessments about Egypt and Syria was twofold. They misjudged the countries' intentions and miscalculated their military capabilities. 
Our coverage of human intelligence, signals intelligence and other sorts was second to none, said Efraim Halevy, a former chief of Mossad, Israel's national intelligence agency. 
We thought we could initially contain any attack or repulse it within a couple of days. We wrongly assessed the capabilities of the Egyptians and the Syrians. In my opinion, that was the crucial failure.
Israel is in a different situation today, Mr. Halevy said. The Syrian armed forces are depleted and focused on fighting their domestic battles, he said. 
The Egyptian Army is busy dealing with its internal turmoil, including a campaign against Islamic militants in Sinai. Hezbollah, the Lebanese militant group, is heavily involved in aiding President Bashar al-Assad of Syria, while the Iranians, 
Mr. Halevy said, are not likely to want to give Israel a reason to strike them, not as the aggressor but as a victim of an Iranian attack. Israel is also much less likely to suffer such a colossal failure in assessment, Mr. Halevy said. 
We have plurality in the intelligence community, and people have learned to speak up, he said. The danger of a mistaken concept is still there, because we are human. But it is much more remote than before.
Many analysts have attributed the failure of 1973 to arrogance. There was a disregarding of intelligence, said Shlomo Avineri, a political scientist at Hebrew University and a director general of Israel's Ministry of Foreign Affairs in the mid-1970s. 
War is a maximization of uncertainties, he said, adding that things never happen the same way twice, and that wars never end the way they are expected to. Like most countries, Israel has been surprised by many events in recent years. 
The two Palestinian uprisings broke out unexpectedly, as did the Arab Spring and the two revolutions in Egypt. In 1973, logic said that Egypt and Syria would not attack, and for good reasons,
said Ephraim Kam, a strategic intelligence expert at the Institute for National Security Studies at Tel Aviv University who served for more than 20 years in military intelligence. 
But there are always things we do not know. Intelligence is always partial, Mr. Kam said, its gaps filled by logic and assessment. The problem, he said, is that you cannot guarantee that the logic will fit with reality.
In his recently published diaries from 1973, Uzi Eilam, a retired general, recalled the sounding of sirens at 2 p.m. on Yom Kippur and his rushing to the war headquarters. 
Eli Zeira passed me, pale-faced, he wrote, referring to the military intelligence chief, and he said: So it is starting after all. They are putting up planes. A fleeting glance told me that this was no longer the Eli Zeira who was so self-assured.

}
\pred{israel, syria, egypt, military, jerusalem}
\true{israel, yom kippur, egypt, syria, military, arab spring}

\doc{Abe's 15-month reversal budget fudges cost of swapping people and butter for concrete and guns. The government of Shinzo Abe has just unveiled its budget for fiscal 2013 starting in April. 
Abe's stated intention was to radically reset spending priorities. He is indeed a man of his word. For this is a budget that is truly awesome for its radical step backward into the past a past where every public 
spending project would do wonders to boost economic growth. It is also a past where a cheaper yen would bring unmitigated benefits to Japan's exporting industries. 
None of it is really true anymore. Public works do indeed do wonders in boosting growth when there is nothing there to begin with. But in a mature and well-developed economy like ours, which is already so well equipped 
with all the necessities of modern life, they can at best have only a one-off effect in creating jobs and demand. And in this globalized day and age, an exporting industry imports almost as much as it exports. 
No longer do we live in a world where a carmaker makes everything within the borderlines of its nationality. Abe's radical reset has just as much to do with philosophy as with timelines. 
Three phrases come to mind as I try to put this budget in a nutshell. They are: from people to concrete,from the regions to the center and from butter to guns. The previous government led by the Democratic Party of Japan 
declared that it would put people before concrete. No more building of ever-empty concert halls and useless multiple amenity centers where nothing ever happens. 
More money would be spent on helping people escape their economic difficulties. They would give more power to the regions so they could decide for themselves what was really good and worked for the local community. 
Guns would most certainly not take precedence over butter. Or rather over the low-fat butter alternatives popular in these more health-conscious times. All of this has been completely reversed in Abe's fiscal 2013 budget. 
Public works spending is scheduled to go up by more than 15 percent while subsistence payments for people on welfare will be thrashed to the tune of more than 7 percent. 
If implemented, this will be the largest cut ever in welfare assistance. The previous government set aside a lump sum to be transferred from the central government's coffers to regional municipalities 
to be spent at their own discretion on local projects. This sum will now be clawed back into the central government's own public works program. The planned increase in spending on guns is admittedly small: 
a 0.8 percent increase over the fiscal 2012 initial budget. It is nonetheless the first increase of its kind in 11 years. And given the thrashing being dealt to welfare spending, the shift in emphasis from butter to guns is clearly apparent. 
One of the Abe government's boasts is that it will manage to hold down the overall size of the budget in comparison with fiscal 2012. The other one is that it will raise more revenues from taxes rather than borrowing. 
True enough on the face of it. But one has to remember the very big supplementary budget that the government intends to push through for the remainder of fiscal 2012. 
The money for that program will come mostly from borrowing. Since the government is talking about a 15-month budget that seamlessly links up the fiscal 2012 supplementary and fiscal 2013 initial budgets, 
they should talk in the same vein about the size of their spending and the borrowing needed to accommodate the whole 15-month package. It will not do to smother the big reset with a big coverup.}

\pred{shinzo abe, japan, budget}
\true{shinzo abe, budget}

\end{document}